\documentclass{article}
\ifdefined\pdfminorversion\pdfminorversion=6\fi

\PassOptionsToPackage{authoryear,round}{natbib}
\usepackage[eandd,final]{neurips_2026}
\usepackage[utf8]{inputenc}
\usepackage[T1]{fontenc}
\usepackage{microtype}

\usepackage{graphicx}
\usepackage{fontawesome5}
\usepackage{tabularx}
\usepackage{colortbl}
\usepackage{multirow}
\usepackage{enumitem}
\usepackage[hidelinks]{hyperref}
\usepackage{url}
\usepackage{xurl}
\usepackage{booktabs}
\usepackage{longtable}
\usepackage{needspace}
\usepackage{array}
\usepackage{amsfonts}
\usepackage{nicefrac}
\usepackage[table]{xcolor}
\usepackage[normalem]{ulem}
\definecolor{revisionblue}{HTML}{145AC4}
\definecolor{revisiongreen}{HTML}{187A3D}
\definecolor{revisiongray}{HTML}{8A8A8A}
\newif\ifshowchanges
\ifdefined\cleanversion\showchangesfalse\else\showchangestrue\fi

\DeclareRobustCommand{\deleted}[1]{\ifshowchanges{\color{revisiongray}\sout{#1}}\allowbreak\hspace{0.25em}\fi}

\usepackage{forest}
\usepackage{tikz}
\usepackage{amssymb}
\usepackage{pifont}
\usepackage{amssymb}
\usepackage[most]{tcolorbox}
\usepackage{xspace}
\usepackage{wrapfig}
\usepackage{subcaption}
\usetikzlibrary{positioning, arrows.meta, shapes.geometric, shapes.misc, fit, calc, shadows, backgrounds}

\definecolor{figurepanelbg}{HTML}{E6DFF3}

\DeclareUnicodeCharacter{02C8}{\textquotesingle}
\DeclareUnicodeCharacter{0251}{a}
\DeclareUnicodeCharacter{02D0}{:}
\DeclareUnicodeCharacter{0259}{e}

\definecolor{successgreen}{HTML}{2E8B57}
\definecolor{failurered}{HTML}{CC4444}
\definecolor{interactionblue}{HTML}{4477AA}
\definecolor{subcatgold}{HTML}{B8860B}
\definecolor{leafgray}{HTML}{555555}
\definecolor{refgray}{HTML}{999999}

\definecolor{tabHead}{HTML}{2C5282}
\definecolor{tabAccent}{HTML}{4477AA}
\definecolor{tabFirstCol}{HTML}{D6E2EE}
\definecolor{tabAlt}{HTML}{EEF2F7}
\definecolor{resourceBg}{HTML}{F5F5F5}
\definecolor{resourceBorder}{HTML}{CBD5E1}
\definecolor{resourceTitleBg}{HTML}{E8EEF4}
\definecolor{resourceTitleText}{HTML}{334B63}

\definecolor{senseRed}{HTML}{C5797E}
\definecolor{thinkGold}{HTML}{D4B14A}
\definecolor{actGreen}{HTML}{A0CC95}
\definecolor{adaptPurple}{HTML}{7491AD}
\definecolor{senseRedTint}{HTML}{F5E6E8}
\definecolor{thinkGoldTint}{HTML}{F8F0D9}
\definecolor{actGreenTint}{HTML}{EDF5EB}
\definecolor{adaptPurpleTint}{HTML}{EAEFF5}
\definecolor{tabHeadDark}{HTML}{2D2D2D}
\definecolor{senseDark}{HTML}{0C447C}
\definecolor{planDark}{HTML}{633806}
\definecolor{actDark}{HTML}{085041}
\definecolor{adaptDark}{HTML}{3C3489}

\newcommand{\sysname}{Act\textperiodcentered\textsc{onomy}\xspace}

\newcommand{\autotraceqda}{\texttt{Automated-Trace-Analysis-Tool}\xspace}

\newcommand{\discoveryqda}{\texttt{LLM-\allowbreak{}powered-\allowbreak{}Discovery-\allowbreak{}Qualitative-\allowbreak{}Analyst}\xspace}
\newcommand{\extensiontool}{\texttt{Automated-Codebook-Extension-Tool}\xspace}

\newcommand{\cat}[1]{\textsc{#1}}
\newcommand{\code}[1]{\textit{#1}}

\title{How to Interpret Agent Behavior}

\author{%
  \parbox{\dimexpr\textwidth-2\tabcolsep\relax}{\centering\normalfont
    {\normalsize\bfseries
    Jie (Sophia)~Gao$^{\spadesuit}$\quad
    Kaiser~Sun$^{\spadesuit}$\quad
    Jen-Tse~Huang$^{\spadesuit}$ \\
    Katherine~Van~Koevering$^{\spadesuit}$\quad
    Sijie~Ji$^{\diamondsuit}$\quad
    Heyuan~Huang$^{\spadesuit}$ \\
    Weiyan~Shi$^{\clubsuit}$\quad
    Zhuoran~Lu$^{\heartsuit}$\thanks{These authors contributed equally.}\quad
    Ziang~Xiao$^{\spadesuit}$\footnotemark[1]\quad
    Daniel~Khashabi$^{\spadesuit}$\footnotemark[1]\quad
    Mark~Dredze$^{\spadesuit}$\footnotemark[1]}\\[0.4em]
    $^{\spadesuit}$The Johns Hopkins University \quad
    $^{\diamondsuit}$California Institute of Technology \\
    $^{\clubsuit}$Northeastern University \quad
    $^{\heartsuit}$The University of Hong Kong\\[0.6em]
    \href{mailto:jgao77@jh.edu}{\texttt{jgao77@jh.edu}}\\
    \begingroup
    \begin{minipage}{\linewidth}
    \centering\normalfont\normalsize\setlength{\parskip}{0pt}
    \Urlmuskip=0mu plus 1mu\relax
    \makebox[1.7em][c]{\raisebox{0pt}[1.1em][0.25em]{\faGlobe}}%
    \textbf{Website:} \textcolor{blue!65!black}{\href{https://actonomy-agents.github.io/}{\nolinkurl{actonomy-agents.github.io}}}\quad
    \makebox[1.7em][c]{\raisebox{-0.15em}[1.1em][0.25em]{\includegraphics[height=1em]{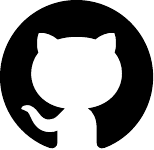}}} 
    \textbf{Code:} \textcolor{blue!65!black}{\href{https://github.com/gaojie058/Act-onomy}{\nolinkurl{gaojie058/Act-onomy}}}\endgraf
    \makebox[1.7em][c]{\raisebox{-0.2em}[1.1em][0.25em]{\includegraphics[height=1.2em]{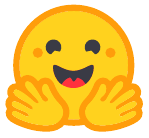}}}%
    \textbf{Dataset:} \textcolor{blue!65!black}{\href{https://huggingface.co/datasets/SophiaGao/Act-ONOMY}{\nolinkurl{SophiaGao/Act-ONOMY}}}\endgraf
    \makebox[1.7em][c]{\raisebox{-0.25em}[1.1em][0.25em]{\includegraphics[height=1.4em]{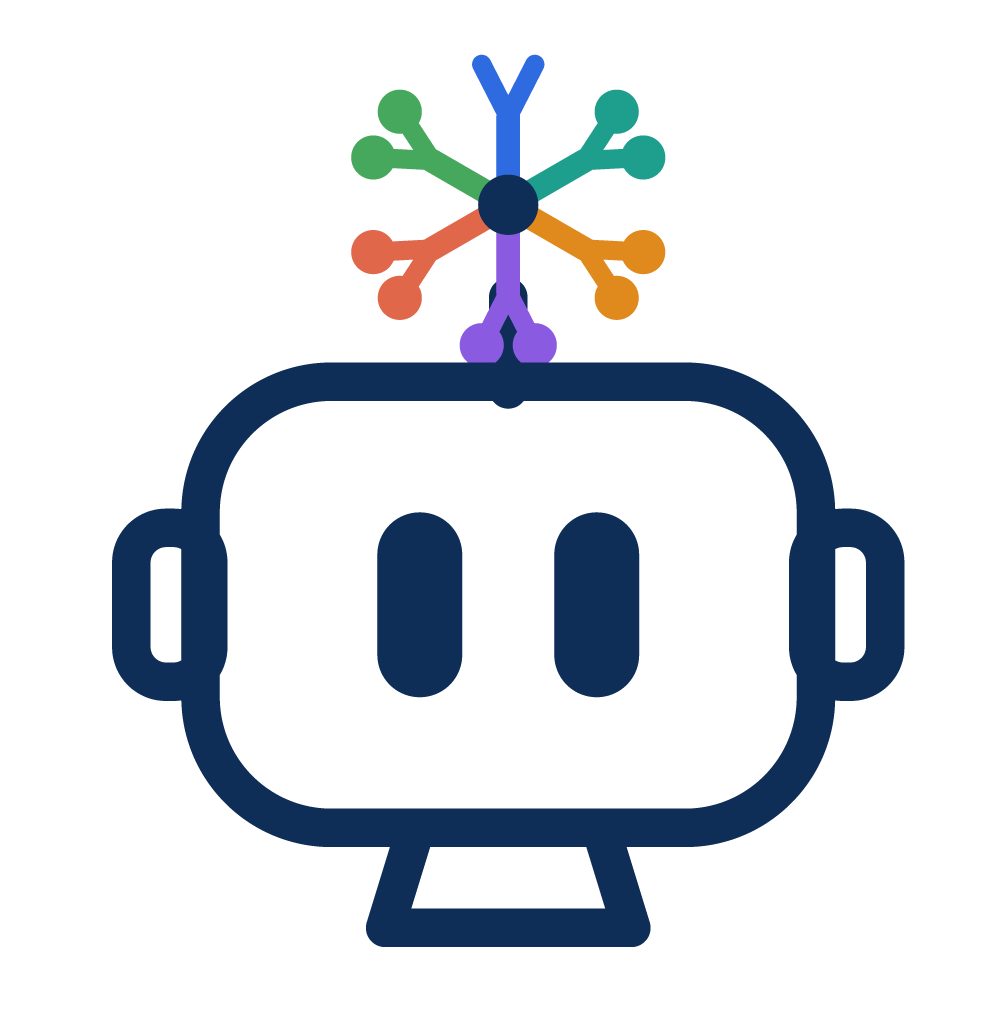}}}%
    \textbf{NeurIPS2026 Workshop:} \textcolor{blue!65!black}{\href{https://iab-agents.github.io/}{\nolinkurl{iab-agents.github.io}}}
    \end{minipage}%
    \endgroup
  }
}

\begin{document}

\maketitle
\setcounter{footnote}{0}

\begin{abstract}
Autonomous agents such as Claude Code and Codex now operate for hours or even days. Understanding their runtime behavior has become critical for downstream tasks such as diagnosing inefficiencies, fixing bugs, and ensuring better oversight.\footnote{By \textit{runtime behavior}, we mean the observable actions an agent takes during execution.} A primary way to gain this understanding is analyzing the reasoning trajectories and execution traces these agents generate. Yet such data remains in unstructured natural-language form, making it difficult for humans to interpret at scale. We introduce \sysname,\footnote{A combination of \textbf{Act}ion and Tax\textbf{onomy}, pronounced /ækˈtɑːnəmi/.} a taxonomy for describing and analyzing agent behavior at runtime. \sysname has two components: (1) \textbf{the taxonomy itself}, developed through Grounded Theory and structured as a three-level hierarchy of 10 actions, 46 subactions, and 120 leaf categories; and (2) \textbf{an open repository} that hosts the living taxonomy, provides an automated analysis pipeline that applies it to agent trajectory analysis, and defines an extension protocol for customization and growth. Our experiments show that \sysname can be used to compare behavioral profiles \textit{across agents} and characterize \textit{a single agent}'s behavior across diverse trajectories, surfacing patterns indicative of failure modes. By providing a shared vocabulary, \sysname aims to help agent researchers, designers, and deployment engineers interpret agent behavior more effectively and consistently, supporting further study of oversight and control.

\end{abstract}

\addtocontents{toc}{\protect\setcounter{tocdepth}{0}}

\vspace{-0.5em}
\begin{flushright}
\textit{``The limits of my language mean the limits of my world.''} \\
\small --- Ludwig Wittgenstein, \textit{Tractatus Logico-Philosophicus}
\end{flushright}
\vspace{-0.75em}

\section{Introduction}

\begin{figure}[!t]
    \centering
    \includegraphics[width=0.96\linewidth]{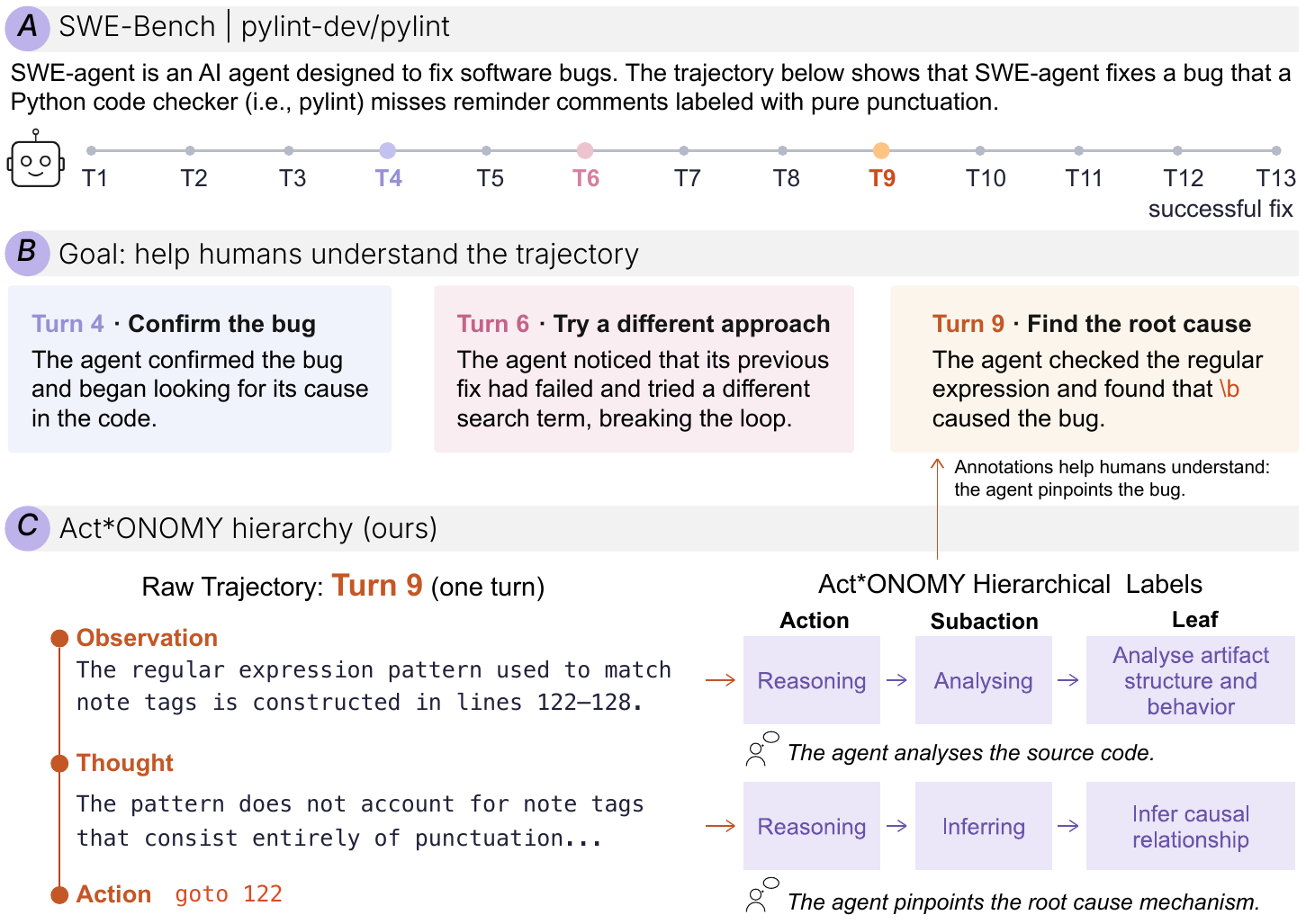}
    \captionsetup{skip=3pt}
    \caption{\sysname supports trajectory interpretation with human-readable action tags. (A) Trajectory overview for pylint-dev/pylint-5859. (B) Three turning points: confirming the bug (Turn~4), trying a different approach (Turn~6), and finding the root cause (Turn~9). (C) Turn~9 spans mapped to action, sub-action, and leafs.}
    \label{fig:motivating_example}
    \vspace{-23pt}
\end{figure}

Modern agents increasingly run autonomously, sometimes for hours or even days~\citep{kwa2026measuringaiabilitycomplete}. They now tackle complex tasks such as solving GitHub issues~\citep{yang2024swe, wang2025openhandsopenplatformai}, navigating web interfaces~\citep{zhou2023webarena}, and conducting research~\citep{lu2024aiscientistfullyautomated}.
Over such extended executions, agents rarely succeed or fail cleanly; more often, they fail and recover repeatedly before reaching a final outcome. Did the agent follow an effective plan or a flawed one? Did it recover from errors, or get stuck in a loop? Did it hallucinate outputs, or ask for help? Understanding what agents actually do during execution is critical for a range of downstream tasks, from diagnosing and repairing agent design bugs and improving runtime efficiency, to building human-centered systems that support meaningful oversight~\citep{rahwan2019machine}.

A central means of developing this understanding is to analyze agent \emph{trajectories}~\citep{cemri2025multiagentllmsystemsfail, yang2024swe}, which are sequential records of an agent's planning, reasoning, and tool use. Recently, trajectory analysis has accordingly emerged as an active research area~\citep{song-etal-2024-trial, xiao2025improving, desmond2025agent}. Traditional analyses rely on quantitative outcome metrics such as task success rate~\citep{yang2024swe}, which reveal \emph{whether} an agent succeeded but little about \emph{how} or \emph{why}~\citep{cemri2025multiagentllmsystemsfail, ma2024agentboardanalyticalevaluationboard}. Without \emph{how} or \emph{why}, it is difficult to identify what to fix, which in turn makes it hard to push success rates higher.
Recent work has therefore turned to more informative qualitative analysis~\citep{cemri2025multiagentllmsystemsfail}, in which humans read trajectories directly to interpret agent behavior and ground subsequent diagnosis or characterization. However, there are two challenges. First, trajectories are unstructured: they appear as long, free-form, and often messy natural-language text, not designed for human consumption~\citep{deshpande2025trail}.
Second, agents are a new kind of artifact that produces behaviors, and the research community has not yet converged on a shared vocabulary for describing what they do~\citep{jia2024can, sreedhar2024simulating}. Researchers studying agent behavior thus lack an established conceptual and vocabulary toolkit to draw on when reporting their findings. Without this toolkit, findings are hard to communicate in ways others can build on, and behavioral knowledge cannot accumulate.
\textbf{The gap between the growing complexity of agent runtime behavior and the vocabulary available to describe and analyze it continues to widen.}
This motivates a fundamental question: \textit{``How do we interpret agent behavior?"}

\begin{figure}[!t]
    \centering
    \includegraphics[width=\linewidth]{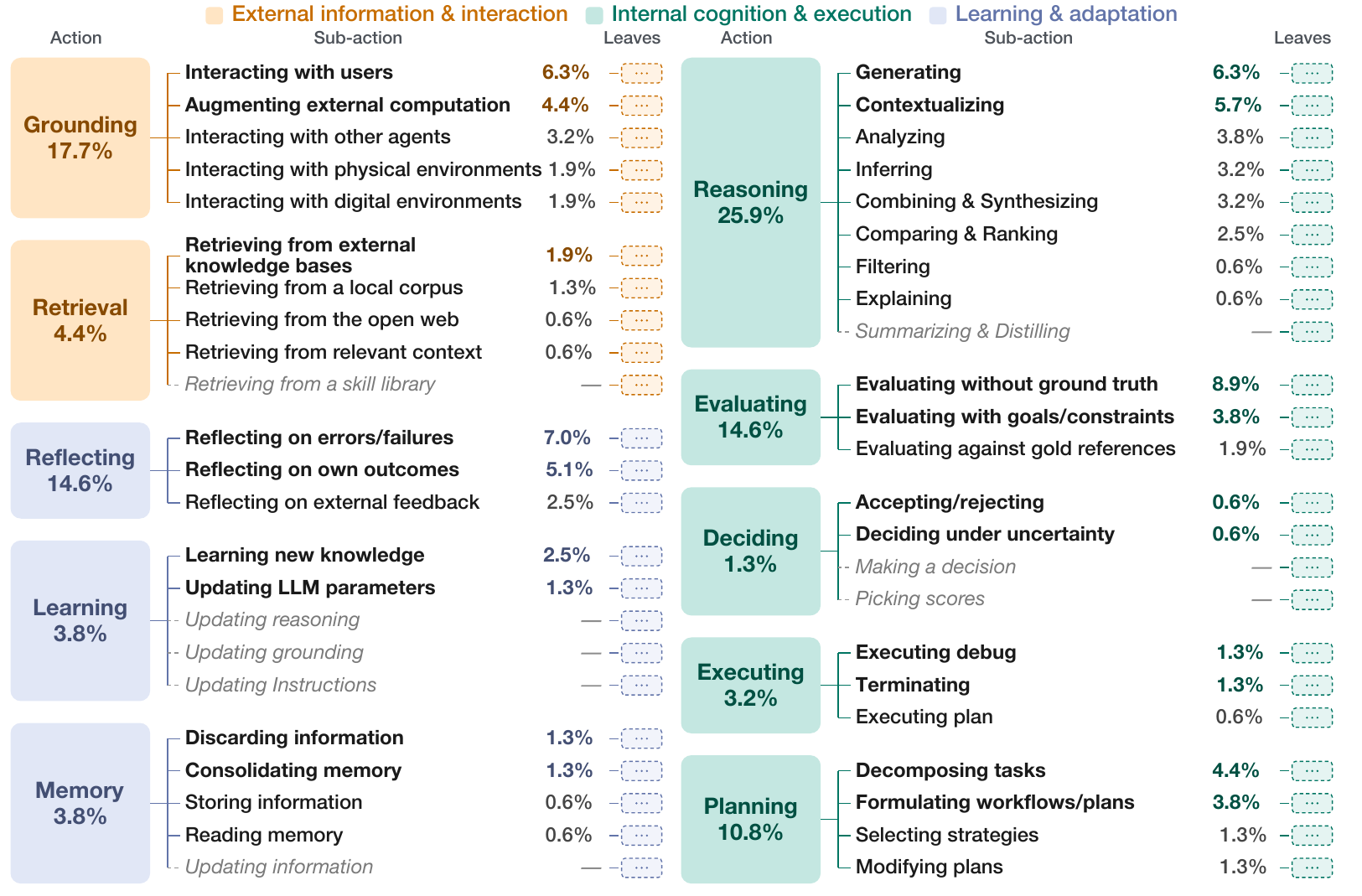}
    \captionsetup{font=small,skip=3pt}
    \caption{\sysname Overview: 10 main actions, 46 subactions, and 120 leaf-level categories. Percentages show the share of the 158 manually selected, paper-grounded behavior-description occurrences in the codebook. Gray rows (``—'') mark theoretically motivated subactions with no sentences in our corpus. }
    \label{fig:actonomy}
     \vspace{-23pt}
\end{figure}

We introduce \sysname, a taxonomy for describing and analyzing agent behavior at runtime. It is grounded in a corpus of 565 behavior descriptions drawn from peer-reviewed publications on AI agents between 2024 and 2026. These descriptions reflect researchers' emerging interpretations of the agent behaviors they have designed or used, expressed in natural language.
We applied a grounded-theory approach~\citep{charmaz2006constructing} in which terms emerged inductively from the corpus, while drawing on existing cognitive architecture frameworks for theoretical grounding~\citep{sumers2024cognitivearchitectureslanguageagents}. It comprises 10 top-level actions (e.g., planning, reasoning), 46 sub-actions (e.g., retrieving from local corpus, executing code), and 120 leaf-categories. Since the field of autonomous agents is evolving rapidly, we envision \sysname as a living taxonomy. We therefore host it as an open GitHub repository, where new behaviors (particularly sub-actions and leaf categories) can be proposed, reviewed, and incorporated by the community. We demonstrate \sysname's potential usage through two case studies: one comparing behavioral profiles across multiple agents, and another characterizing a single agent's behavior within diverse trajectories. In summary, our main contributions are as follows:

\begin{itemize}[nosep,leftmargin=1.2em]
\item \textbf{Taxonomy.} We propose \sysname, the first hierarchical taxonomy for describing and analyzing observable agent behavior. It comprises 10 actions, 46 sub-actions, and 120 leaf categories, theoretically grounded in literature~\citep{sumers2024cognitivearchitectureslanguageagents} and empirically grounded in 565 agent behavior descriptions drawn from the latest research.
\item \textbf{Automatic analysis tool.} We provide an \autotraceqda{} pipeline that automates the application of \sysname for agent trajectory analysis, enabling users to build behavioral profiles of agents at scale.
\item \textbf{Extensibility.} We host an open repository to maintain \sysname as a living taxonomy that absorbs new sub-actions as the technology evolves. We also provide an automatic tool that lets users adapt the taxonomy according to their preferences and domain requirements.
\item \textbf{Use cases.} We demonstrate \sysname's utility through two case studies, applying it to agent trajectories to compare behavior \textit{across} and \textit{within} agents.
\end{itemize}

\par\vspace{4pt}
\noindent
\begin{minipage}[b]{\linewidth}
\section{\sysname: Describing and Analyzing Agent Behaviors at Scale}
\label{sec:framework}
\vspace{-5.5pt}
\noindent
\begin{minipage}[t]{0.50\linewidth}
\vspace{0pt}
To build a shared vocabulary and conceptual framework for describing and analyzing agent behavior, we constructed \sysname through a grounded theory approach. Because the taxonomy is our primary contribution, we present it first and detail our methodology in \S\ref{sec:method}.

\subsection{Taxonomy Overview}

\sysname organizes agent behaviors into three broad categories: external information and interaction, internal cognition and execution, and learning and adaptation (Figure~\ref{fig:actonomy}).

\textbf{The first cluster captures how the agent acquires information and interacts with the external world.} \cat{\textbf{Grounding}} describes behaviors

\end{minipage}\hfill
\begin{minipage}[t]{0.48\linewidth}
\vspace{0pt}
\centering
\includegraphics[width=0.98\linewidth,trim=0 17bp 0 11bp,clip]{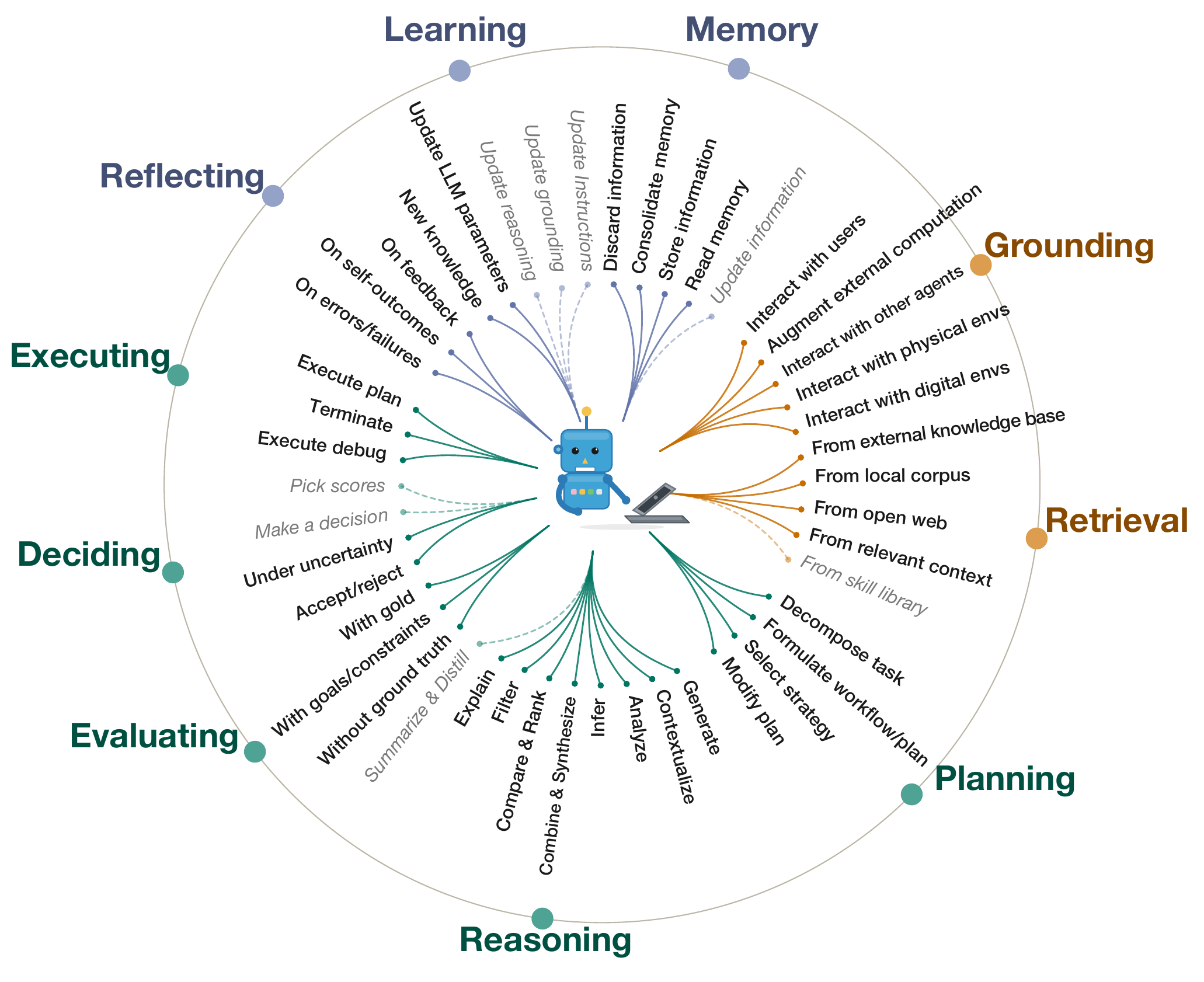}
\captionsetup{font=small,skip=1pt}
\captionof{figure}{An at-a-glance map of \sysname.}
\label{fig:action-space}
\end{minipage}
\par\vspace{0pt}
\end{minipage}
\par

through which the agent exchanges information with external entities (e.g., users, physical or digital environments, peer agents, and external tools or computational resources). \cat{\textbf{Retrieval}} describes behaviors through which the agent obtains task-relevant information from various information sources, including external knowledge bases, local corpora, and the open web.

\textbf{The second cluster captures how agents conduct internal cognition and carry out concrete execution.} \cat{\textbf{Reasoning}} refers to the internal cognitive operations the agent performs to produce new content. Notably, reasoning was predominantly used to describe operations over information already present in the agent's context; it does not reach outside the agent. It is also the most frequently described behavior in our corpus (25.9\%), encompassing operations such as generating, analyzing, inferring, comparing \& ranking \& filtering, contextualizing, and combining \& synthesizing (e.g., \textit{``combine information from multiple sources...to produce a coherent solution''} (P2)).  Although often co-occurring with reasoning, \cat{\textbf{Planning}} (10.8\%) plays a distinct role: it is primarily used to decide what to do next, including decomposing tasks into subgoals, formulating workflows, selecting strategies, and modifying plans. For example, \textit{``the planner accurately interprets user intent and formulates a comprehensive analysis workflow''} (P1). \cat{\textbf{Evaluating}} captures actions in which the agent judges quality or correctness, either against gold-standard results or against specific criteria such as goals, requirements, constraints, rubrics, or domain rules. For example, \textit{``independently checks whether objectives are truly complete, preventing the orchestrator from advancing when the main agent incorrectly believes a task is finished''} (P25). It also frequently describes situations in which the agent must evaluate in the absence of ground truth, relying instead on internal standards and heuristics (e.g., \textit{LLM-as-a-judge scoring}, \textit{visual} or \textit{behavioral correctness checks}).
We treat \cat{\textbf{Deciding}} as a distinct action because it marks an important ``commitment" phase in the overall task. For instance, selecting among options surfaced by upstream actions, or determining whether to engage at all. Similarly, we include \cat{\textbf{Executing}} as a distinct action because it captures phases in which the agent commits to and carries out an action, e.g., executing a plan or terminating the run by delivering a final answer. Unlike cognitive actions such as planning or reasoning, \cat{\textbf{Executing}} refers only to the act of carrying out, not to the deliberation that precedes it. For example, \textit{an insertion agent executes this plan for HLS-C optimization} (P29). 

\textbf{The third cluster captures how the agent learns from experience and adapts to real-world complexity.} \cat{\textbf{Reflecting}} describes actions in which the agent examines its own process, including reflecting on failures, reflecting on self-generated outcomes, and incorporating external feedback. Notably, in our corpus, reflection is most often used to describe ``thinking about" past behavior rather than ``fixing" it. For example, \textit{``Given the [Chat History] REFLECT carefully on the AI assistant's last response''} (P25). In contrast, \cat{\textbf{Learning}} has less empirical grounding in our corpus and reflects a more theory-driven framing; following \citet{sumers2024cognitivearchitectureslanguageagents}, we define it as the process of updating an agent's knowledge, reasoning procedures, or parameters in ways that persist across episodes. However, in our corpus, \cat{\textbf{Learning}} and \cat{\textbf{Reflecting}} are often used interchangeably; we nevertheless retain them as separate categories to mark this subtle but meaningful distinction. \cat{\textbf{Memory}} captures actions that operate on the agent's external memory resources, e.g., memory banks, scratchpads, to-do lists. The actions include storing, updating, and discarding information.

\begingroup
\setlength{\columnsep}{5pt}
\subsection[Distribution and Co-occurrence of Reported Agent Behaviors]{\texorpdfstring{\parbox[t]{0.55\textwidth}{Distribution and Co-occurrence\\of Reported Agent Behaviors}}{Distribution and Co-occurrence of Reported Agent Behaviors}}

\label{sec:auto-analysis}
\begin{wrapfigure}[13]{r}{0.42\textwidth}
    \vspace{\dimexpr-\intextsep-3\baselineskip\relax}
    \centering
    \includegraphics[width=\linewidth]{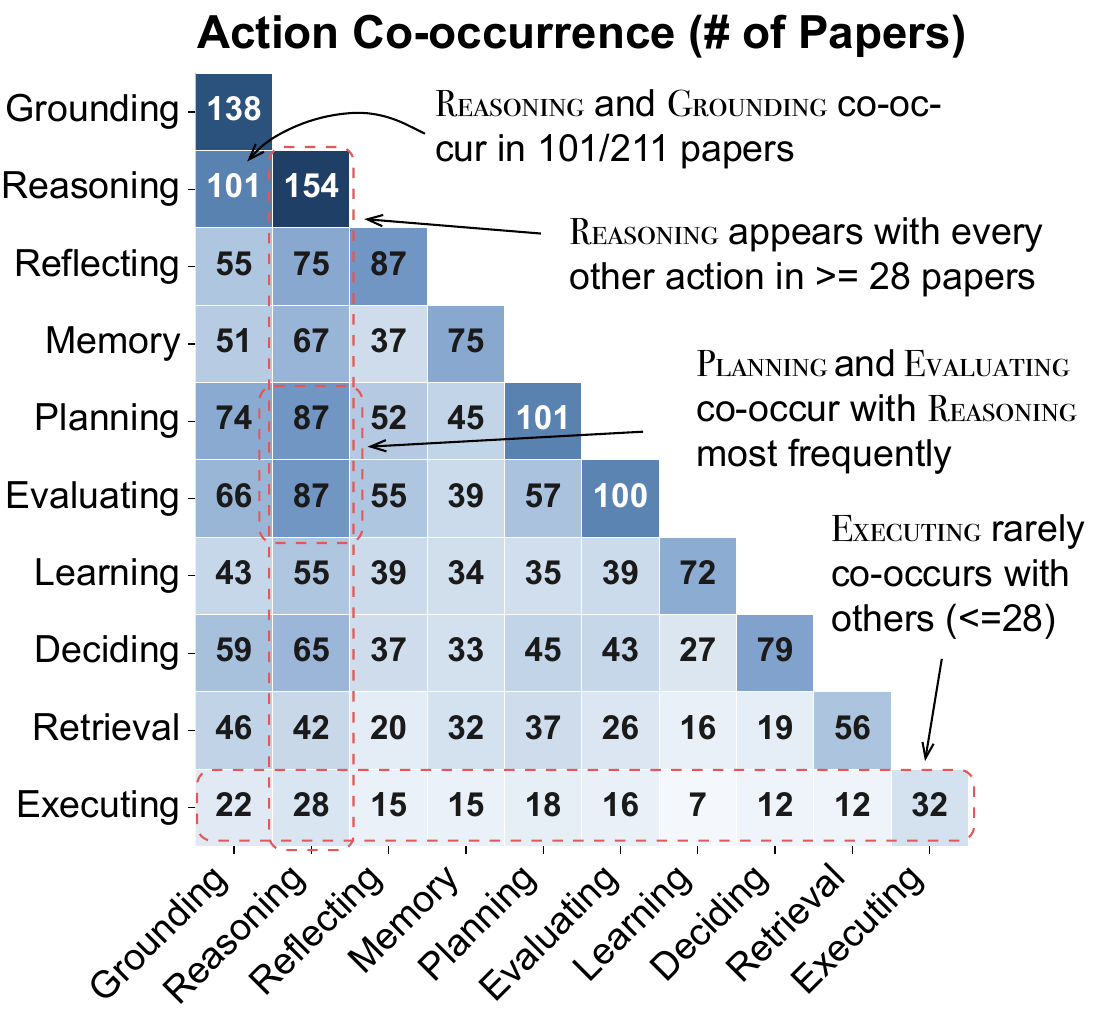}
    \caption{Action co-occurrence.}
    \label{fig:large-scale-analysis}
\end{wrapfigure}

We use \sysname to examine how behavioral descriptions are distributed across a diverse set of agent papers. Our dataset comprises 3,455 behavioral descriptions extracted from 211 behavior-related agent papers curated by the \texttt{awesome-language-agents} GitHub list,\footnote{\href{https://github.com/ysymyth/awesome-language-agents}{\nolinkurl{github.com/ysymyth/awesome-language-agents}}; data extracted in May 2026.} spanning safety, evaluation, software engineering, computer use, and web automation.  We found that, \textbf{(1) Frequency across taxonomy levels:} At the Action level, \cat{Grounding} (21.0\%) and \cat{Reasoning} (20.2\%) are the most frequently reported categories, followed by \cat{Reflecting} (10.8\%), \cat{Memory} (10.4\%), \cat{Planning} (10.2\%), and \cat{Evaluating} (10.1\%). At the Sub-action level, Evaluating without ground truth (8.0\%) and Contextualizing (7.8\%) are the most frequent categories. At the Leaf level, packaging prior reasoning (5.9\%) and self-correcting a step (4.4\%) are the most frequent categories (see Appendix~\ref{fig:appendix:distribution_scale_combined}a). \textbf{(2) \cat{Reasoning} co-occurs broadly with other actions} (Figure~\ref{fig:large-scale-analysis}): it appears with every other action in $\geq$28 papers and pairs with \cat{Grounding} in 101 papers, the most frequent pair across the corpus. \cat{Executing}, by contrast, co-occurs sparsely ($\leq$28), indicating that execution is less often explicitly described alongside other actions in this corpus.  This suggests that \cat{Reasoning} is relatively well established and widely accepted action.
More details are provided in Appendix~\ref{sec:appendix:behavior_scale}.
\par
\endgroup

\section{Construct and Extend \sysname: A Grounded Theory Approach}
\label{sec:method}

\begin{figure}[!hb]
    \centering
    \includegraphics[width=\linewidth]{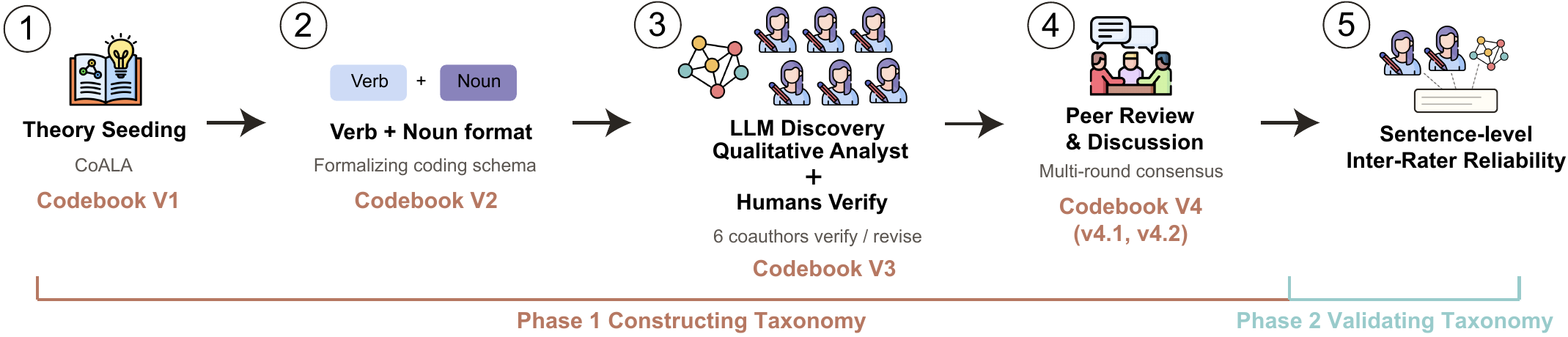}
    \caption{An Overview of our Grounded Theory \citep{charmaz2006constructing} pipeline to construct \sysname.}
\label{fig:method}
\vspace{-10pt}
\end{figure}

We construct \sysname via a grounded theory approach~\citep{charmaz2015grounded, charmaz2006constructing}, a qualitative method well-suited for surfacing vocabulary in emerging, ill-defined domains and previously used to study agent failure modes~\citep{cemri2025multiagentllmsystemsfail}. We further anchor it in the Cognitive Architectures for Language Agents (CoALA) framework~\citep{sumers2024cognitivearchitectureslanguageagents}, which provides foundational action-space definitions such as planning, reasoning, and learning. Figure~\ref{fig:method} overviews our method.

\paragraph{Phase 1: Constructing Taxonomy.}\label{sec:method-phase1}
We build \sysname from 565 behavioral descriptions (20 construction papers). The initial pool contained 780 sentences (35 papers), with 664 sentences (28 papers) assigned to the construction set and 116 sentences (7 papers) held out for codebook testing. Six co-authors manually reviewed the construction set and removed 99 sentences (8 off-topic papers), leaving the 565 descriptions used to construct \sysname (Appendix~\ref{sec:data-collection}).
\textbf{(1) Seed (V1$\to$V2).} We initialized \textit{Codebook V1} based on the guiding theoretical framework~\citep{sumers2024cognitivearchitectureslanguageagents}, including high-level categories (e.g., planning, reasoning, retrieval) along with descriptive sentences characterizing agent behavior. We then reformulate each seed behavior into a ``verb + noun'' form (\textit{V2}) to ensure consistent wording and a uniform level of abstraction for taxonomy building.\footnote{Each ``verb + noun'' description is a qualitative ``code,'' and the resulting set of codes forms the codebook.}
\textbf{(2) Scale and review (V2$\to$V3).} After building the theoretical seed codebook, we expanded the initial codebook. This is done through examination of 565 candidate behavioral descriptions for relevance to the scope of \sysname and their categories. This is done with the assistance of \discoveryqda{} (Appendix~\ref{app:discovery-judge}), which provides preliminary assessments, based on the given behavior description.  And then six co-authors reviewed the descriptions and made the final inclusion decisions, they independently \emph{verify}, \emph{accept}, or \emph{propose} a new code, \emph{rename} for clarity, or \emph{discard} it for every (\textit{description}, \textit{suggested code}, \textit{fit or not}) pair. 
\textbf{(3) Finalize (V3$\to$V4.2).} After human manual review, the codebook contained additional evidence from the codebook. The six co-authors collectively reviewed and discussed V3 over multiple rounds, producing V4.1 with 146 leaf categories and 195 descriptions; V4.2 reorganized it into 120 leaf categories and 158 descriptions (Table~\ref{tab:codebook-evolution}).

\paragraph{Phase 2: Validating the Taxonomy.}\label{sec:phase2} We test the V4.2 codebook on the held-out set. Two authors first used 30 held-out sentences to identify and discuss ambiguous definitions in V4.2 and clarify code descriptions. With the clarified codebook fixed, the two authors independently coded 20 unseen sentences (Cohen's $\kappa=0.87/0.72$ at the action/sub-action levels); agreement between the \textit{LLM-powered deductive analyst} and the primary coder was $\kappa=0.74/0.71$. We then apply the finalized V4.2 codebook to all sentence descriptions in the held-out set using the same \textit{LLM-powered deductive analyst}: no new actions emerge, and any new subactions are minor variations of existing ones, indicating that \sysname{} has reached initial saturation at the action level and is close to saturation at the sub-action level (Table~\ref{tab:codebook-evolution}). Full saturation at the leaf level was not expected. Overall, the construction and validation process required substantial human effort: all six co-authors contributed over 6 hours of annotation, with the primary and secondary annotators investing more than 20 and 10 hours, respectively.

\paragraph{Toolkit for Extending \sysname.}
We treat \sysname as a living taxonomy: the main actions and sub-actions are expected to remain relatively stable, while the leaf level continues to expand as new agents emerge and new behavior descriptions are added. We support extension through \extensiontool{}, which automatically propagates codebook changes to dependent files (Appendix~\ref{sec:extension}).

\WFclear
\renewcommand{\autotraceqda}{\texttt{Automated-\allowbreak{}Trace-\allowbreak{}Analysis-\allowbreak{}Tool}\xspace}
\renewcommand{\extensiontool}{\texttt{Automated-\allowbreak{}Codebook-\allowbreak{}Extension-\allowbreak{}Tool}\xspace}
\section{How Can \sysname Support Agent Behavior Analysis?}
\label{sec:case-study} 

A primary purpose of \sysname{} is to support downstream tasks such as trajectory analysis for understanding agent behaviors. We first propose an automated pipeline for describing and analyzing trajectories, and then present two case studies that illustrate its use in practice.

\subsection{Toolkit for Applying \sysname in Agent Behavior Analysis at Scale.}
\label{sec:trace-analysis-auto}

We leverage LLM-powered qualitative coding to deductively apply a codebook for downstream trajectory analysis. We develop \autotraceqda (hereafter, the tool), which performs the following six steps:
\textbf{(1) Preprocessing.} Given an agent trajectory, the tool parses it into a sequence of per-turn triples of \emph{observation}, \emph{thought}, and \emph{action}. \textbf{(2) Behavioral indicator extraction.} Within each turn, the tool identifies behavior-indicating spans in both the \textit{thought} and the \textit{action}. \textbf{(3) Codebook assignment.} Each extracted span is annotated with an \textit{action--subaction--leaf} label. When no suitable subaction or leaf exists, the tool proposes new ones to extend the codebook. \textbf{(4) Ambiguous case resolution.} Since trajectories are expressed in natural language, they contain many ambiguous cases that might prevent LLMs from assigning labels. We curated an ambiguity case library containing ambiguous quotes, human-adjudicated labels, and decision rules, which LLMs can consult when they encounter ambiguous cases. Humans can refine the library when needed. \textbf{(5) Aggregation and summarization.} The tool computes statistics over the annotated trajectory, segments it into coherent sessions, and generates a natural-language summary describing what the agent did in each session. \textbf{(6) Profile presentation.} The statistics, summaries, and grounded annotations are compiled into a behavioral profile that users can interactively inspect to understand the agent's behavior. 
We implement this pipeline as a reusable skill and evaluate its annotation reliability using three models in their respective agent harnesses (Appendices~\ref{sec:appendix:skill} and~\ref{sec:additional-validation}).

\subsection{Understanding Behavior Distributions \textit{Across Agents}}
\label{sec:case-study-1}

One potential use case for \sysname is to characterize each agent's behavior profile both qualitatively and quantitatively. We selected three agents from different domains and tasks for automatic trajectory analysis using \autotraceqda: AG2~\citep{wu2023autogenenablingnextgenllm}, HyperAgent~\citep{phan2025hyperagentgeneralistsoftwareengineering}, and SWE-Agent~\citep{yang2024swe}. We collected 100 publicly available traces released from MAST~\citep{cemri2025multiagentllmsystemsfail} for each agent  \footnote{Raw trajectories: \url{https://github.com/multi-agent-systems-failure-taxonomy/MAST/tree/main/traces/AG2/experiments}.}.

Overall, the three agent trajectory samples show a similar pattern: \cat{Reasoning} and \cat{Executing} are the most common actions, while \cat{Learning} is the least common.

\textbf{AG2} is a general-purpose framework for building multi-agent systems that work together through conversation. The 100 AG2/MathChat trajectories we sampled cover GSM-Plus math word problems (see \S\ref{sec:raw-math-trajectories}). 
The \sysname annotation shows that these sampled trajectories have higher proportions of \cat{Evaluating} (13.6\%), \cat{Grounding} (18.3\%), and \cat{Deciding} (2.8\%) actions and lower proportions of \cat{Retrieval} (1.5\%) and \cat{Reflecting} (0.8\%) actions than the trajectories sampled from the other two agents. The higher share of \cat{Evaluating} may reflect both the math tasks and the verification instructions in the trajectories.

\textbf{HyperAgent} is a multi-agent system for software engineering tasks. The 100 sampled trajectories record HyperAgent's attempts to resolve GitHub issues from SWE-bench Lite. These trajectories were released through MAST (see \S\ref{sec:raw-math-trajectories}). Within these sampled trajectories, \cat{Reasoning} is the most common action (24.0\%), followed by \cat{Executing} (17.3\%). The share of \cat{Executing} actions is lower than in the AG2 (20.3\%) and SWE-Agent (29.5\%) samples. \cat{Reflecting} actions also account for a higher proportion of annotated quotes in the HyperAgent sample (4.7\%) than in the AG2 (0.8\%) and SWE-Agent (3.7\%) samples.

\textbf{SWE-Agent} is also a software engineering agent for resolving GitHub issues. The 100 trajectories we sampled involve resolving repository-level issues, including bug fixes and feature enhancements. Compared with HyperAgent, the SWE-Agent sample has a higher proportion of \cat{Executing} actions (29.5\% vs.\ 17.3\%) and a lower proportion of \cat{Reasoning} actions (15.7\% vs.\ 24.0\%). Thus, \cat{Executing} is the most common action in the SWE-Agent sample, whereas \cat{Reasoning} is the most common in the HyperAgent sample. The SWE-Agent sample also has smaller shares of \cat{Grounding} (10.9\% vs.\ 12.9\%) and \cat{Reflecting} (3.7\% vs.\ 4.7\%) actions. Although both systems address software engineering tasks, differences in their workflows and sampled issues may shift the balance between recorded reasoning and tool execution.

\textbf{More importantly, \sysname also describes actions beyond the categories identified in the authors' original analysis.} For instance, our analysis identified non-trivial proportions of \cat{Planning}, \cat{Evaluating}, \cat{Deciding}, \cat{Reflecting}, and \cat{Memory} actions in SWE-Agent's trajectories, whereas the original action-pattern analysis~\citep[Table~8]{yang2024swe} grouped frequent command sequences into \emph{Reproduction}, \emph{Localization (File)}, \emph{Localization (Line)}, \emph{Editing}, and \emph{Submission}. These categories mainly summarize what commands the agent executes and what task those commands serve. Our annotations also describe how the agent plans its next steps, evaluates results, and revises its approach around these operations.

\textbf{Assessing Reproducibility.}
To assess the reproducibility of the behavioral profiles, we tested their stability across different trajectory samples and the consistency of action rankings across three annotator models (Appendix~\ref{sec:profile-validation}).

\begin{figure}[!t]
    \centering
    \includegraphics[width=\linewidth]{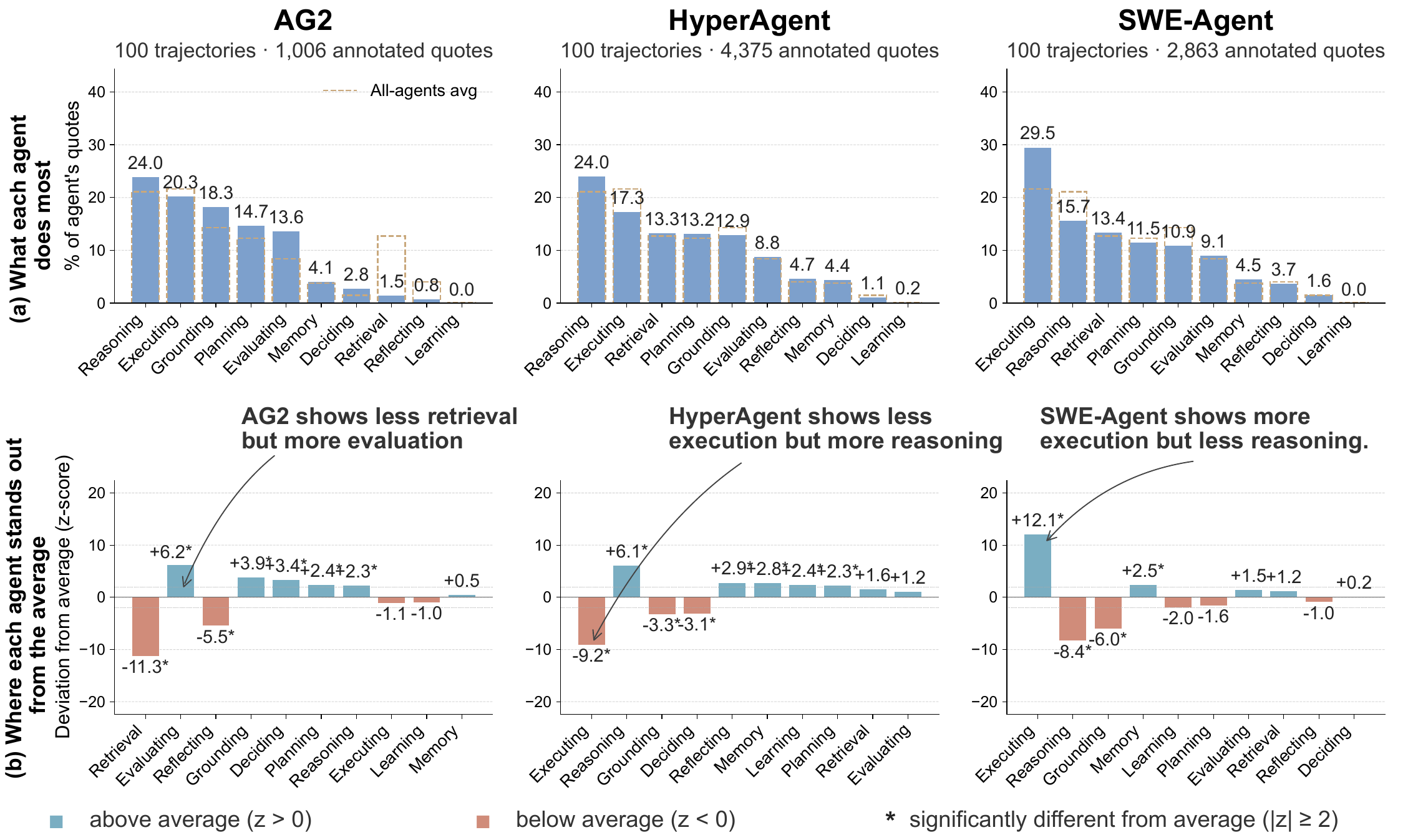}
    \caption{Three agents show distinct behavioral profiles. (a)~Action distribution for each agent; (b)~each agent's deviations from the cross-agent average. }
    \label{fig:case-study1}
\end{figure}

\subsection{Understanding Behavior Distributions \textit{Within a Single Agent}.}

\label{sec:case-study-2}
\begin{figure}[!t]
    \centering
    \includegraphics[width=\linewidth]{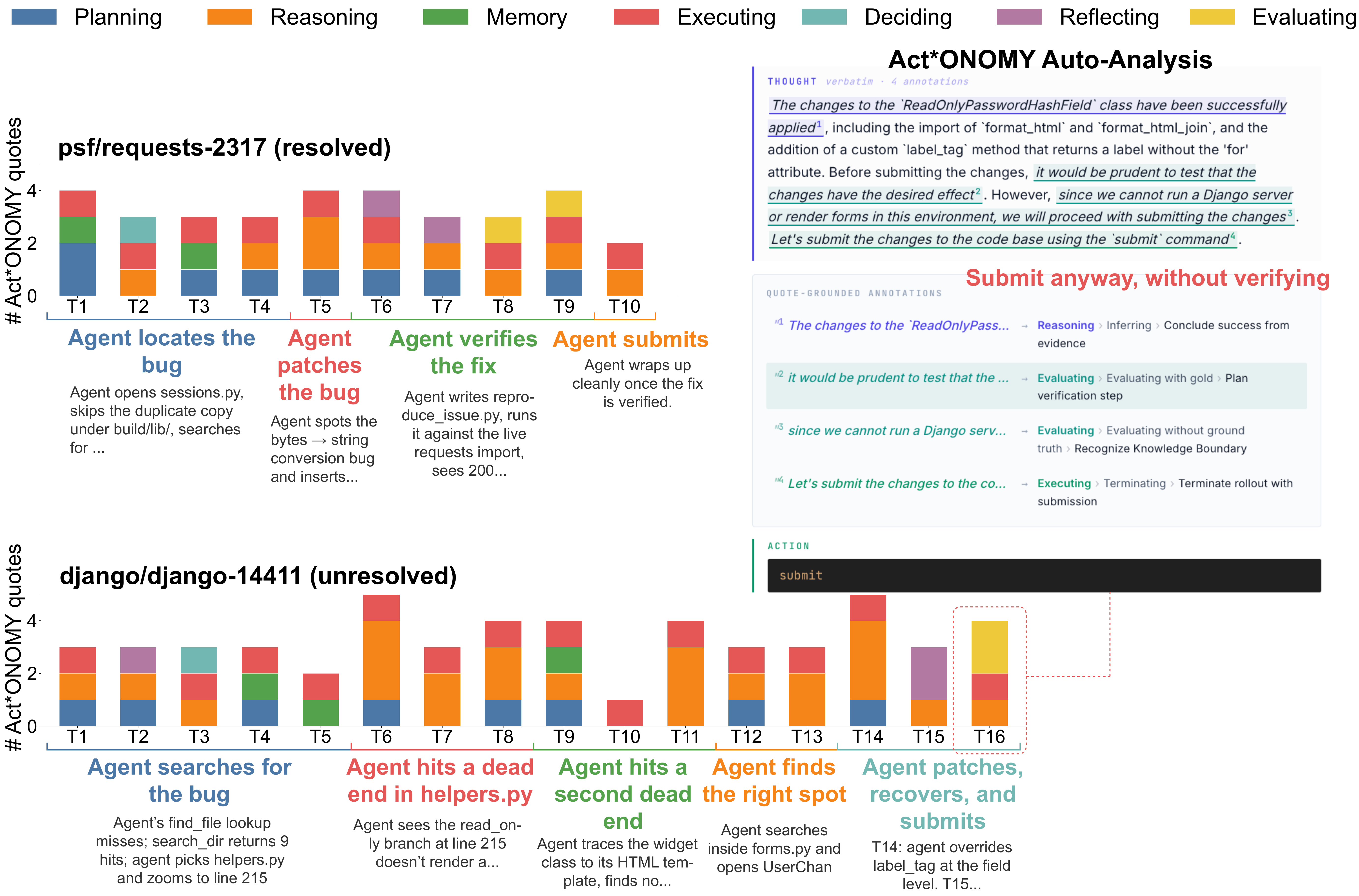}
    \caption{\textbf{Two SWE-agent trajectories produce contrasting behavioral shapes.} Stacked bars show per-turn \sysname categories assigned by \autotraceqda, accompanied by its automatically generated natural-language session summaries. The callout zooms in on the leaf-level, quote-grounded labels that pinpoint specific behaviors accompanying the agent's decision.}
    \label{fig:case-study-2}
\end{figure}

After examining how \sysname differentiates \emph{across} agents, we now turn to how it characterizes variation \emph{within} a single agent's behavior. We select two trajectories generated by SWE-agent~\citep{yang2024swe} on GitHub issue-repair tasks from SWE-bench: Trace~1, \texttt{psf/requests-2317}, which the agent resolved, and Trace~2, \texttt{django/django-14411}, which the agent did not resolve. Figure~\ref{fig:case-study-2} presents the per-turn breakdown of both runs, the \autotraceqda-generated \sysname tags, and the session-level summaries characterizing the agent's behavior. Each tag is a quote-grounded annotation containing three hierarchical levels: Action, Sub-action, and Leaf. We count each annotation once and aggregate tags by their top-level Action in the distributions below.

\textbf{Trajectory-level shape.} Although both trajectories are dominated by \cat{Reasoning} and \cat{Executing}, a pattern consistent with SWE-agent's software engineering task, they diverge in their behavioral composition. Trace~1 includes 10 turns and 33 \sysname tags, organized into four phases: \emph{locate the bug}, \emph{patch the bug}, \emph{verify the fix}, and \emph{submit}. Trace~2, by contrast, has 16 turns and 53 tags across five phases: \emph{search for the bug}, \emph{hit a dead end}, \emph{hit a second dead end}, \emph{find the correct file}, and \emph{patch, recover, and submit}. The two runs also differ in their internal balance: Trace~1 distributes tags evenly between \cat{Reasoning} and \cat{Executing} (N=9 each), whereas Trace~2 is heavily skewed toward \cat{Reasoning} (22 of 53 tags) over \cat{Executing} (14 tags). {}
The reason is because Trace~2 has more search and dead-end phases, meaning a more complex problem-solving process. 
These pattern differences could explain the final outcomes: Trace~1 successfully resolves the task so tages are distributed evenly, while Trace~2 is a failed running where the agent needs to figure out the reason behind each dead end. 

\textbf{Surfacing a fine-grained candidate failure pattern.} Beyond high-level differences in action distributions, leaf-level analysis reveals subtle yet critical insights into the agent's problem-solving process. As illustrated on the right of Figure~\ref{fig:case-study-2}, \autotraceqda assigns a series of quote-grounded labels to the agent's thought and action: 
\begin{itemize}
    \item \code{Reasoning $\to$ Inferring $\to$ Conclude success from evidence} (``\textit{The changes ... have been successfully applied}''); 
    \item \code{Evaluating $\to$ Evaluating with gold $\to$ Plan verification step} (``\textit{it would be prudent to test that the changes have the desired effect}''); 
    \item \code{Evaluating $\to$ Evaluating without ground truth $\to$ Recognize knowledge boundary} (``\textit{since we cannot run a Django server ... we will proceed with submitting}''); 
    \item \code{Executing $\to$ Terminating $\to$ Terminate rollout with submission} (``\textit{Let's submit the changes ... using the submit command}''). 
\end{itemize}
Taken together, these four labels indicate a \emph{``submit anyway, without verifying''} failure pattern: the agent acknowledges the need for verification, recognizes its inability to perform one, and proceeds to submit nonetheless. Trace~2 also shows an increase in \cat{Reflecting} tags immediately before submission (T15), whereas Trace~1's \cat{Reflecting} tags appear during verification (T6--T7).

\paragraph{Pattern Recurrence Check.} Will this pattern recur in a larger sample? We extended the analysis to 32 public SWE-agent trajectories, balanced between 16 resolved and 16 unresolved tasks, with 2,289 tags. Using 10,000 trajectory-level bootstrap resamples, we estimated reflection's share in the final three turns at $0.095$ for unresolved runs and $0.028$ for resolved runs (difference $0.067$, 95\% CI $0.012,0.122$). Recognizing a knowledge boundary followed by submission occurred in 4/16 unresolved and 1/16 resolved runs. These findings provide additional support for the use of \autotraceqda in discovering recurring behavioral patterns (Appendix~\ref{sec:pattern-validation}).

\section{Related Work}
\label{sec:related-work}

\paragraph{Trajectory analysis of LLM agents.}
Most analyses of LLM agents rely on quantitative outcome metrics: AgentBench~\citep{liu2025agentbenchevaluatingllmsagents}, AgentBoard~\citep{ma2024agentboardanalyticalevaluationboard}, and SWE-bench~\citep{jimenez2024swebenchlanguagemodelsresolve} report task success, progress, or issue-resolution rates. Such metrics tell us \emph{whether} an agent succeeded but little about \emph{how} or \emph{why}. Analyzing trajectories can be valuable to understand the dynamics of agent behavior, however, agent trajectories are long, free-form, often-messy natural-language text not designed for human consumption, which makes systematic manual analysis costly and hard to scale. Recent work that does inspect trajectories mainly focuses on failure modes or ad-hoc analysis. Their focus is mainly on \emph{failure}: \citet{cemri2025multiagentllmsystemsfail} taxonomize multi-agent failure modes, and \citet{kapoor2024agents} catalog reliability gaps, leaving a spectrum of agent behaviors largely unexamined. A few exceptions are agent-specific instead, e.g., manual analyses of how SWE-agent navigates repositories~\citep{yang2024swe}, with categories tailored to their study aims. These analyses motivate our shared vocabulary for characterizing agent runtime behavior across systems. \sysname targets this gap with a descriptive taxonomy that describes and analyzes agent behavior, paired with an automated pipeline that automatically turns unstructured trajectories into human-readable behavioral profiles at scale.

\paragraph{Action-space and cognitive-architecture frameworks.}
A complementary line of work conceptualizes agents through cognitive architectures. CoALA~\citep{sumers2024cognitivearchitectureslanguageagents} organizes language agents around external actions (interacting with users, environments, and tools) and internal actions (retrieval, reasoning, learning). Earlier theoretical foundations from Newell's unified theory of cognition~\citep{newell1994unified, anderson2003newell, vernon2016desiderata} offer operational criteria for cognition such as adaptivity, robustness, and self-awareness, which map naturally onto behaviors observable in LLM agents. More practically, WorldAPIs~\citep{ou2025worldapisworldworthapis} approaches the action space empirically by inducing primitive APIs from wikiHow tutorials. At the level of work activities, \citet{cai2026where} develop a hierarchical ontology from O*NET to analyze where AI software and robotic systems are used. 
These frameworks provide a conceptual basis for the action space.
\sysname complements them with the descriptive vocabulary needed for trajectory analysis: 10 actions, 46 sub-actions, theoretically grounded in cognitive-architecture theory and empirically grounded in behavioral descriptions written by AI researchers.

\paragraph{Qualitative coding with LLMs.}
Qualitative methods such as grounded theory~\citep{charmaz2006constructing, charmaz2015grounded} and thematic analysis~\citep{braun2006using, clarke2017thematic} have a long tradition of turning unstructured natural-language data into a structured, human-interpretable vocabulary. Recent work shows that LLMs can scale parts of this pipeline~\citep{fischer-biemann-2024-exploring, gao2024collabcoder, gao2025efficiencyrigortrustworthyllmpowered, parfenova-etal-2025-text}. We build on this line in two ways. First, we use human qualitative coding to derive \sysname inductively from agent papers while drawing on cognitive-architecture theory. Second, we provide \autotraceqda, an LLM-as-qualitative-coder pipeline that applies \sysname to agent trajectories at scale.

\section{Discussion}

\paragraph{Behavioral interpretability as a complement to mechanistic interpretability.}
Most current work on understanding language-model systems looks \emph{inside} the model via probing~\citep{belinkov2022probing}, sparse autoencoders~\citep{cunningham2023sparse}, and circuit analysis~\citep{wang2022interpretability,conmy2023automated}. \sysname argues for a complementary lens that looks \emph{at the trajectory}: agents are now complex enough to warrant ethological description~\citep{rahwan2019machine}, not only mechanistic dissection. Outcome metrics such as pass@1 or turn count~\citep{kapoor2024agents,ma2024agentboardanalyticalevaluationboard} cannot distinguish a clean \emph{locate--patch--verify--submit} run from a \emph{search--dead-end--submit-anyway} run, whereas a behavioral profile, with its quote-grounded leaf categories, makes this distinction explicit. A natural next step is to link behavioral codes back to internal model state---for instance, probing for \cat{Reflecting}~\citep{shinn2023reflexion} or \cat{Planning}~\citep{yao2023react,zhou2024lats}---so that \sysname can serve as a bridge between behavioral and mechanistic interpretability.

\paragraph{Implications for agent observability and oversight.}
Behavioral profiles open a practical surface for agent monitoring~\citep{desmond2025agent,deshpande2025trail}. Across-agent profiles surface \emph{architectural fingerprints} that distinguish single-task math agents from repository-scale multi-agent systems (Case Study~\ref{sec:case-study-1}); within-agent profiles surface \emph{trajectory-level patterns} and \emph{candidate failure-associated patterns}, such as the elevated share of \cat{Reasoning} on complex tasks and the ``\emph{submit anyway, without verifying}'' pattern surfaced by quote-grounded T16 labels (Case Study~\ref{sec:case-study-2}). Ad-hoc behavioral analyses already appear across recent agent papers~\citep{yang2024swe,cemri2025multiagentllmsystemsfail}, but each invents its own categories. \sysname consolidates them into a shared codebook, providing a basis for evaluating downstream uses: behavioral regression testing across agent versions~\citep{ribeiro2020checklist,bhardwaj2026agentassay,rehan2026tdad,ma2025rethinking}, behavioral drift detection in production~\citep{qin2026implicitmembenchmeasuringunconsciousbehavioral}, and oversight~\citep{bowman2022measuring} of long-running agents whose raw traces would otherwise be too voluminous to read~\citep{xiao2025improving}.  By giving researchers, designers, and end users a shared vocabulary, \sysname aims to make agent behavior something that can be discussed, compared, and built upon rather than re-described from scratch by every new system.

\section{Conclusion}

As modern AI agents grow increasingly complex and autonomous, describing and analyzing their behavior has become correspondingly difficult. We began with a fundamental question: \emph{how do we interpret agent behavior?} In response, we introduce \sysname, a taxonomy that provides a shared vocabulary for agent behavior, empirically grounded in 565 behavioral descriptions retained from 20 construction papers within an initial pool of 35 agent papers (2024--2026) and theoretically anchored in cognitive-architecture research. \sysname comprises two components: (1)~\textbf{the taxonomy itself}, organized as a three-level hierarchy of 10 actions, 46 subactions, and 120 leaf categories; and (2)~\textbf{an open repository} that hosts the living taxonomy alongside two supporting artifacts: \autotraceqda, an automated analysis pipeline that produces quote-grounded annotations of raw trajectories, and an extension protocol that enables the community to incorporate new actions over time. Two case studies illustrate the application of \sysname for both across-agent and within-agent analysis. The validation checks support action-level profile comparisons in the sampled settings, while fine-grained interpretation and diagnostic use require further validation. We position \sysname as a starting point: a living vocabulary for the agents we study today, designed to grow into one for the agents we have yet to build.

\begin{ack}
We thank the organizers of the first Interpreting Agent Behavior (IAB) Workshop, including Teresa Yeo and Boyuan Zheng, for their support. This work was supported by Schmidt Sciences. Jie Gao is supported by the Malone Postdoc Fellowship. Katherine Van Koevering is supported by the DSAI Postdoc Fellowship. Daniel Khashabi is supported by ONR award (N00014-24-1-2089). Kaiser Sun was supported in part by DARPA (SciFY). The U.S. Government is authorized to reproduce and distribute reprints for Governmental purposes. The views and conclusions contained in this publication are those of the authors and should not be interpreted as representing official policies or endorsements of DARPA or the U.S. Government.
\end{ack}

\bibliographystyle{abbrvnat}
\bibliography{references}

\addtocontents{toc}{\protect\setcounter{tocdepth}{2}}
\clearpage
\appendix

\tableofcontents
\clearpage

\section{Limitations}
\label{sec:limitations}

\sysname has several limitations. First, our scope is manually constructed on 565 behavioral sentences retained from 20 construction papers within a 35-paper initial pool (Appendix~\ref{sec:data-collection}), and the preliminary coding saturation we report (\S\ref{sec:phase2}) is within this corpus. We tested the generalizability of \sysname on a large-scale dataset in \S\ref{sec:auto-analysis} and \S\ref{sec:appendix:behavior_scale}: action and sub-action levels remain relatively stable, while new behavioral descriptions continue to emerge at the leaf level, which our extension protocol is designed to absorb. Second, the codebook is built from descriptions of agent behavior in research papers. These descriptions may emphasize what agents are designed to do (e.g., ``the agent reflects'') and omit low-level behaviors that become apparent only when examining execution traces. As an initial check, we compare the coverage of our codebook with that of a codebook derived from execution traces. Third, the construction process relied on Claude Code, which may have introduced bias. We then used several models to extract behaviors and apply the codebook and assessed agreement for their code assignments.

\section{Broader Impacts}
\label{sec:impacts}

\sysname is a descriptive tool meant to make agent runtime behavior easier to describe and compare for researchers, designers, and end users, and to give failure-mode taxonomies such as MAST~\citep{cemri2025multiagentllmsystemsfail} a vocabulary for what the agent was doing before it failed. We flag four considerations for responsible use. First, a \sysname profile is an interpretation of what an agent did on the trajectories we can observe, not a ground-truth account of what agents actually do; when deciding whether to ship an agent, profiles should be read alongside the raw trajectories they summarize, with the trajectories themselves remaining the primary evidence. Second, a widely adopted vocabulary can in turn constrain how researchers describe behaviors that do not yet fit existing codes, so our extension protocol (Section~\ref{sec:extension}) and versioned releases keep the codebook open to revision, and downstream users should contribute new codes. Third, the construction corpus is drawn from NeurIPS, ICML, ICLR, and the ACL Anthology (2024--2026), so behaviors documented in regional venues, industry reports, or non-English literature are under-represented. Finally, \autotraceqda{} relies on a commercial LLM API, and as commercial models are retired, results from earlier model versions also become harder to reproduce; the taxonomy and extension protocol are model-agnostic, and we release the skills and codebook so that other sufficiently capable open-weight models can implement \autotraceqda{}.

\section{Corpus and Annotation Details}
\label{sec:data-collection}

\paragraph{Corpus collection.} \textbf{Paper selection.} We collected a corpus of \textbf{35 peer-reviewed agent papers} (P1--P35) from NeurIPS, ICML, ICLR, and the ACL Anthology (2024--2026), filtered by the keywords \emph{behavior analysis} or \emph{analysis} in the title or abstract; workshop papers were excluded. The corpus deliberately spans diverse agent-research subdomains (LLM-agent benchmarks and evaluation, multi-agent systems, embodied agents, reinforcement-learning theory, and a range of domain applications such as software engineering, medical AI, and hardware design). We split the 35 papers in advance into a 28-paper construction set and a 7-paper held-out set, with the latter reserved for the reliability check (Section~\ref{sec:phase2}).
\textbf{Behavior-description extraction.} Using the \discoveryqda{} (Appendix~\ref{app:discovery-judge}, behavior extraction), we extracted \textbf{780 behavior-description sentences} from the 35 papers. After author review of the extractions against their source papers, 8 construction papers were judged off-topic for an agent-behavior taxonomy (position pieces, theory papers without runtime behavior, or systems whose described actions do not generalize) and their 99 sentences were removed. The remaining 565 sentences from the 20 incorporated construction papers were used to construct \sysname; the 116 sentences from the 7 held-out papers were reserved for the reliability check (Table~\ref{tab:corpus-splits}). The final corpus is released in our GitHub repository.

\begin{table}[!htbp]
\centering
\footnotesize
\renewcommand{\arraystretch}{1.2}
\setlength{\tabcolsep}{6pt}
\caption{Corpus split.}
\label{tab:corpus-splits}
\vspace{4pt}
\begin{tabularx}{\linewidth}{@{}l r r >{\raggedright\arraybackslash}X@{}}
\toprule
\textbf{Subset} & \textbf{Papers} & \textbf{Sentences} & \textbf{Role} \\
\midrule
Construction (incorporated) & 20 & 565 & Six co-authors reviewed all 565 sentences; the released V4.2 codebook contains 120 leaf categories. \\
Construction (not incorporated) & \phantom{0}8 & \phantom{0}99 & All 99 sentences reviewed, but the papers were judged off-topic for an agent-behavior taxonomy, so their sentences were dropped. \\
Held-out (reliability check) & \phantom{0}7 & 116  & Reserved before initial construction. Of 50 sentences sampled for the human check, 30 informed V4.2 refinement and 20 remained unseen until independent evaluation (Section~\ref{sec:agreement-procedure}). \\
\midrule
\textbf{Total} & \textbf{35} & \textbf{780} & \\
\bottomrule
\end{tabularx}
\end{table}

\paragraph{Annotation process.} We assigned codes to the 565 behavioral descriptions in two steps: an LLM proposed a candidate code for each description, and six co-authors reviewed every (sentence, suggested-code) pair. \textbf{Step 1: \discoveryqda{} suggests a code.} For each behavioral description, the \discoveryqda{} (Appendix~\ref{app:discovery-judge}) compared the description against the current codebook and either matched it to an existing code or proposed a new candidate, in each case with a verbatim quote as evidence. \textbf{Step 2: co-authors review each pair.} Six co-authors split the corpus into batches of 4--5 papers each, with one as the main reviewer. For every (sentence, suggested-code) pair, the reviewer first checked whether the sentence was in scope, verified the quoted span against the source paper, and then \emph{accepted}, \emph{renamed}, \emph{proposed} a new code, or \emph{discarded} the sentence. All six co-authors contributed at least 6 hours; the primary and secondary annotators spent over 20 and 10 hours respectively. Per-coauthor annotation instructions are released in our GitHub repository.

\paragraph{Agreement Measurement.}\label{sec:agreement-procedure}
Two authors first coded 30 held-out sentences under V4.1 in two batches of 15, independently within each batch. The first batch was a warm-up followed by discussion. Before discussing the second batch, Cohen's $\kappa$ was $0.53$ for actions and $0.48$ for sub-actions. About 1.5 hours of discussion identified ambiguous definitions and motivated explicit boundary rules, producing V4.2. The authors then independently coded the remaining 20 previously unseen sentences under the fixed V4.2, with no discussion during coding. 
Agreement was calculated before adjudication: $\kappa=0.87$ for actions and $0.72$ for sub-actions. 

\begingroup
\footnotesize
\renewcommand{\arraystretch}{1.15}
\setlength{\tabcolsep}{3pt}
\newlength{\corpusTableWidth}
\setlength{\corpusTableWidth}{\dimexpr\linewidth-8\tabcolsep\relax}
\begin{longtable}{@{}
  >{\raggedright\arraybackslash}p{0.05\corpusTableWidth}
  >{\raggedright\arraybackslash}p{0.18\corpusTableWidth}
  >{\raggedright\arraybackslash}p{0.20\corpusTableWidth}
  >{\raggedright\arraybackslash}p{0.46\corpusTableWidth}
  >{\raggedleft\arraybackslash}p{0.11\corpusTableWidth}@{}}
\caption{The 20 incorporated construction papers, ordered by paper ID. \emph{Sentences} is the per-paper count of behavior-description sentences extracted by the \discoveryqda{} and kept after author review (565 in total).}
\label{tab:incorporated-papers}\\
\toprule
\textbf{ID} & \textbf{Cite} & \textbf{Domain} & \textbf{Title} & \textbf{Sentences} \\
\midrule
\endfirsthead
\multicolumn{5}{@{}l}{\tablename~\thetable{} (continued)}\\
\toprule
\textbf{ID} & \textbf{Cite} & \textbf{Domain} & \textbf{Title} & \textbf{Sentences} \\
\midrule
\endhead
\midrule
\multicolumn{5}{r@{}}{Continued on next page}\\
\endfoot
\bottomrule
\endlastfoot
P1  & \citet{xiao2024cellagentllmdrivenmultiagentframework}  & LLM Agents                         & CellAgent: LLM-Driven Multi-Agent Framework for Natural Language-Based Single-Cell Analysis & 30 \\
P2  & \citet{paul2026benchmarkdeepinformationsynthesis}  & LLM Agents (Benchmark)             & A Benchmark for Deep Information Synthesis                                                & 26 \\
P3  & \citet{chong2026talkevaluatediagnoseuseraware}  & LLM Agents (Evaluation)            & Talk, Evaluate, Diagnose: User-aware Agent Evaluation with Automated Error Analysis        & 28 \\
P4  & \citet{jiang2026medvrannotationfreemedicalvisual}  & Medical AI                         & MedVR: Annotation-Free Medical Visual Reasoning via Agentic Reinforcement Learning         & 28 \\
P8  & \citet{yang2026metrostrategyinductionexpert}  & LLM Agents (Dialogue)              & METRO: Towards Strategy Induction from Expert Dialogue Transcripts for Non-collaborative Dialogues & 28 \\
P9  & \citet{hu2025reprobenchagenticaisystems}  & LLM Agents (Benchmark)             & REPRO-Bench: Can Agentic AI Systems Assess the Reproducibility of Social Science Research? & 28 \\
P11 & \citet{li2026humanmachinepreliminaryturing} & Speech Interaction                 & Human or Machine? A Preliminary Turing Test for Speech-to-Speech Interaction               & 28 \\
P13 & \citet{qin2026implicitmembenchmeasuringunconsciousbehavioral} & LLM Behavior (Benchmark)           & ImplicitMemBench: Measuring Unconscious Behavioral Adaptation in Large Language Models     & 27 \\
P14 & \citet{wang2026gameplayqabenchmarkingframeworkdecisiondense} & Multimodal Video Understanding     & GameplayQA: A Benchmarking Framework for Decision-Dense POV-Synced Multi-Video Understanding of 3D Virtual Agents & 28 \\
P16 & \citet{li2026agencybenchbenchmarkingfrontiersautonomous} & LLM Agents (Benchmark)             & AgencyBench: Benchmarking the Frontiers of Autonomous Agents in 1M-Token Real-World Contexts & 30 \\
P17 & \citet{meyerson2025positionscalingllmagents} & Position Paper                     & Position: Scaling LLM Agents Requires Asymptotic Analysis with LLM Primitives              & 30 \\
P21 & \citet{liao2025wordsimplyopinionreader} & Multi-Agent Social Simulation      & My Words Imply Your Opinion: Reader Agent-Based Propagation Enhancement for Personalized Implicit Emotion Analysis & 25 \\
P22 & \citet{liu-etal-2025-tell} & LLM Agents (Safety)                & Tell Me What You Don't Know: Enhancing Refusal Capabilities of Role-Playing Agents via Representation Space Analysis and Editing & 28 \\
P27 & \citet{yi2025autotascalableautomatedthematic} & Multi-Agent Systems                & Auto-TA: Towards Scalable Automated Thematic Analysis via Multi-Agent Large Language Models with Reinforcement Learning & 26 \\
P28 & \citet{liu2026e2edevbenchmarkinglargelanguage} & Software Engineering               & E2Edev: Benchmarking Large Language Models in End-to-End Software Development Task         & 35 \\
P29 & \citet{li2026chathlssystematicdesignautomation} & Hardware Design / EDA              & ChatHLS: Towards Systematic Design Automation and Optimization for High-Level Synthesis    & 30 \\
P31 & \citet{desanti2024geometricactiveexplorationmarkov} & Reinforcement Learning Theory      & Geometric Active Exploration in Markov Decision Processes: the Benefit of Abstraction      & 27 \\
P32 & \citet{hu2024infiagentdabenchevaluatingagentsdata} & LLM Agents (Benchmark)             & InfiAgent-DABench: Evaluating Agents on Data Analysis Tasks                                & 28 \\
P33 & \citet{triantafyllou2024agentspecificeffectscausaleffect} & Reinforcement Learning Theory      & Agent-Specific Effects: A Causal Effect Propagation Analysis in Multi-Agent MDPs           & 26 \\
P34 & \citet{shi-etal-2024-opex} & Embodied Agents                    & OPEx: A Component-Wise Analysis of LLM-Centric Agents in Embodied Instruction Following    & 29 \\
\midrule
\multicolumn{4}{r}{\textbf{Total}} & \textbf{565} \\
\end{longtable}
\endgroup

\section{Discovery Qualitative Analyst}
\label{app:discovery-judge}

We developed the \discoveryqda{} to handle parts of the pipeline that are too tedious to do by hand. It has two main operations: identifying behavioral descriptions in a corpus paper, and, given a description, either matching it to an existing code in the current codebook or proposing a new one. We apply it at four points throughout this work:

\paragraph{Behavior extraction.} Given a corpus paper, the \discoveryqda extracts candidate behavior-description sentences, that is, sentences in which the authors describe what their agent does at runtime.

\paragraph{Code suggestion (V2~$\to$~V3).} Given the current codebook and a behavior description, the \discoveryqda proposes either an existing code that fits or a new candidate code; the six co-authors then decide whether to accept it.

\paragraph{Code assignment for inter-rater reliability.} Given Codebook V4 and a behavior description from the held-out set, the \discoveryqda{} assigns a code to the behavior description, used to compute the human--LLM Cohen's $\kappa$. 
\paragraph{Saturation test.} Following the agreement check, we apply \discoveryqda{} to a held-out set and test whether there are new codes that emerge (\S\ref{sec:phase2}).

All four share the same model and pipeline, differing only in the instruction block, the artifacts loaded into context (codebook version, behavior description), and the output format. We implement the \discoveryqda{} as an agent skill and release it in our GitHub repository.

\section{Codebook Evolution}
\label{app:codebook-evolution}

We summarize how the codebook evolved alongside the annotation process described in Section~\ref{sec:data-collection}. After V3, no new top-level actions are added and later versions only restructure; V4.2 consolidates the hierarchy into 10 actions, 46 sub-actions, and 120 leaves. 

\begin{table}[h]
\centering
\footnotesize
\renewcommand{\arraystretch}{1.25}
\setlength{\tabcolsep}{3pt}
\caption{Overview of how \sysname codebook evolved across iterative annotation process.}
\label{tab:codebook-evolution}
\vspace{4pt}
\begingroup
\newcommand{\verpill}[2]{%
  \tcbox[on line, colback=#1, colframe=#1, boxrule=0pt, arc=2pt, boxsep=0pt,
         left=3pt, right=3pt, top=0.6pt, bottom=0.6pt]%
  {\textcolor{white}{\textbf{\textsc{#2}}}}}
\newcommand{\phasetag}[2]{\,\textit{\scriptsize\textcolor{#1}{#2}}}
\resizebox{\linewidth}{!}{%
\begin{tabular}{l r r r >{\raggedright\arraybackslash}p{0.46\linewidth}}
\specialrule{1pt}{0pt}{0pt}
\rowcolor{tabHead}
\textcolor{white}{\textbf{Version}} &
\textcolor{white}{\textbf{Actions}} &
\textcolor{white}{\textbf{Sub-actions}} &
\textcolor{white}{\textbf{Leaf Categories}} &
\textcolor{white}{\textbf{Key change}} \\
\specialrule{1pt}{0pt}{0pt}
\verpill{leafgray}{V1}\phasetag{leafgray}{seed}    & 6  & 17 & 53  & Bullet-list seed drawn from CoALA~\citep{sumers2024cognitivearchitectureslanguageagents}. \\
\rowcolor{tabAlt}
\verpill{leafgray}{V2}\phasetag{leafgray}{normalize} & 6  & 6  & 35  & Action labels normalized to Verb+Noun form with plain-English explanations.  \\
\verpill{thinkGold}{V3}\phasetag{thinkGold}{expand}  & 13 & 13 & 113 & 664 behavior-description sentences from 28 construction papers (Appendix~\ref{sec:data-collection}) were collected through the \discoveryqda{} and reviewed by six co-authors. Seven new top-level categories appear: \cat{Planning}, \cat{Reflection}, \cat{Tool Use}, \cat{Synthesis}, \cat{Evaluate}, \cat{Boundary-Aware}, \cat{Role Conditioning}. \\
\rowcolor{adaptPurpleTint}
\verpill{adaptPurple}{V4.1}\phasetag{adaptPurple}{restructure} & 11 & 60 & 146 & Three-level tree (Action $\to$ Sub-action $\to$ Leaf Category) with paper-anchored examples, with an \emph{Unclassified} bin for items pending placement. \\
\rowcolor{actGreenTint}
\verpill{actGreen}{V4.2}\phasetag{actGreen}{final $\bigstar$} & \textbf{10} & \textbf{46} & \textbf{120} & Final reorganization: the \emph{Unclassified} bin is classified into mainline categories; redundant leaves and actions are merged or renamed; low-frequency codes are consolidated or removed; \cat{Generating} becomes a sub-action under \cat{Reasoning}. \\
\specialrule{1pt}{0pt}{0pt}
\end{tabular}%
}
\endgroup
\end{table}

\section{Large-Scale Analysis of Agent Behavioral Descriptions}
\label{sec:appendix:behavior_scale}

To test how \sysname generalizes beyond the 20-paper incorporated construction corpus, we applied it to a much larger and noisier set of 211 papers and 3{,}455 behavioral descriptions. This appendix records how that corpus was assembled.

\paragraph{Paper collection.} We started from the \texttt{awesome-language-agents} GitHub list,\footnote{\href{https://github.com/ysymyth/awesome-language-agents}{\nolinkurl{github.com/ysymyth/awesome-language-agents}}} a community-maintained index of language-agent research that catalogues several hundred papers across safety, evaluation, software engineering, computer use, web automation, and embodied tasks. We pulled the list as of April 2026, downloaded every paper reachable from it, removed duplicates and broken links, and kept entries whose title or abstract indicated that the paper analyzes agent runtime behavior or reports per-step traces of an agent solving a task. This left 211 papers, intentionally broader and noisier than our 20-paper incorporated construction corpus, to examine applicability across a broader literature sample.

\paragraph{Behavioral description extraction.} We ran \discoveryqda{} (Appendix~\ref{app:discovery-judge}, behavior extraction) once per paper, keeping only its verbatim-quote hallucination guard and skipping the per-sentence author review used in the construction run. This produced 3{,}455 behavioral descriptions, roughly 16 per paper on average.

\paragraph{Codebook annotation.} For each of the 3{,}455 sentences, \discoveryqda{} (Appendix~\ref{app:discovery-judge}, code suggestion) suggested an Action, Sub-action, and Leaf-category label, and proposed a new code whenever no existing code fit. The resulting label distributions is listed in Figure~\ref{fig:large-scale-analysis}. 

\begin{figure}[h]
    \centering

    \makebox[\linewidth][c]{%
    \begin{subfigure}[t]{\linewidth}
        \centering
        \includegraphics[width=\linewidth]{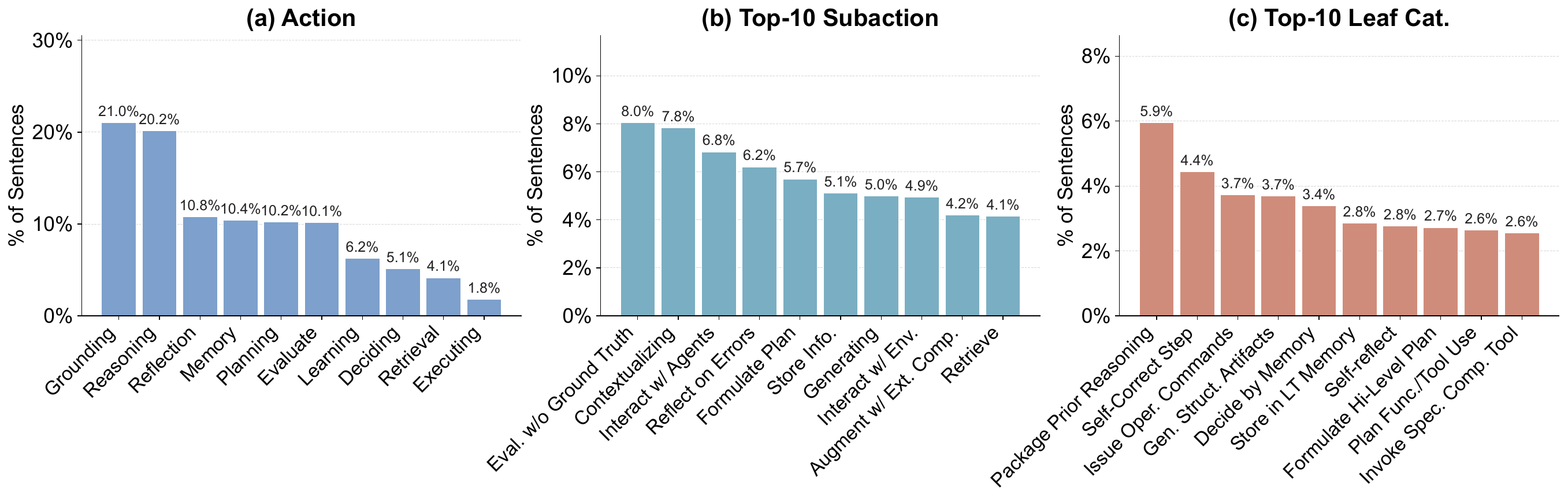}
        \caption{Frequency of each code across the 3{,}455 sentences, shown at the Action, Sub-action, and Leaf levels (top-10 for Sub-action and Leaf).}
        \label{fig:appendix:distribution_scale_bar}
    \end{subfigure}}

    \vspace{6pt}

    \makebox[\linewidth][c]{%
    \begin{subfigure}[t]{\linewidth}
        \centering
        \includegraphics[width=\linewidth]{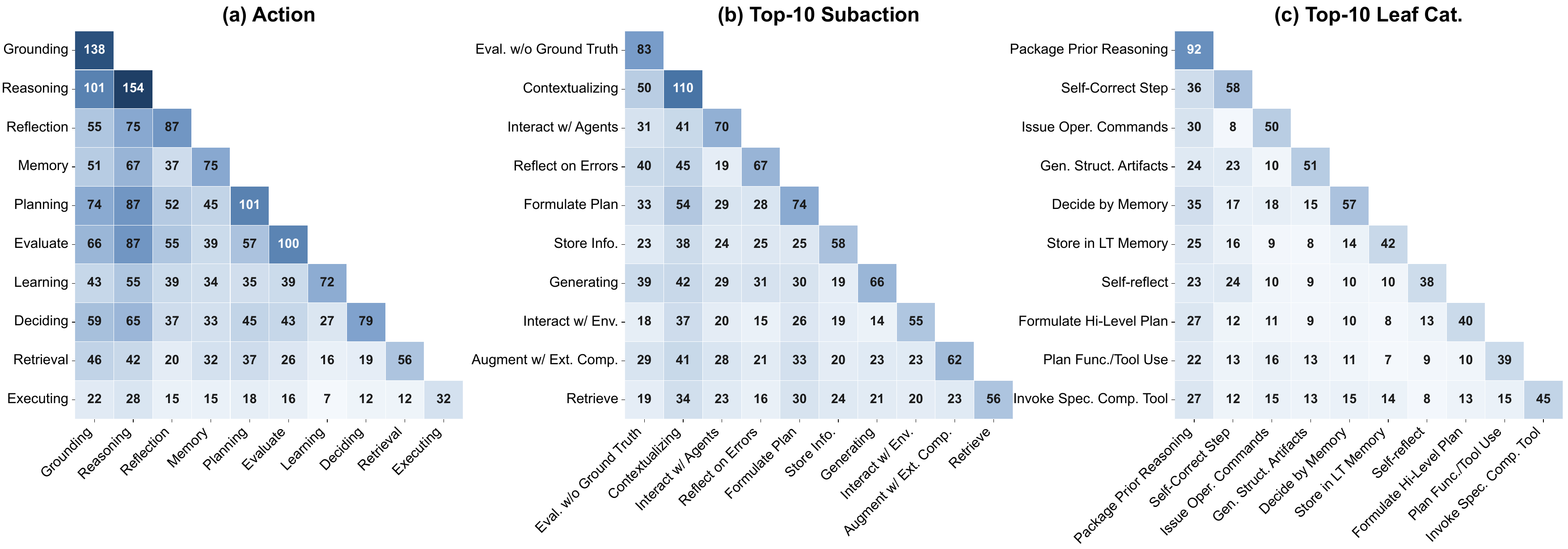}
        \caption{Per-paper co-occurrence of codes at the Action, Sub-action, and Leaf levels; each cell counts how many of the 211 papers contain at least one sentence labeled with both codes.}
        \label{fig:appendix:distribution_scale_heatmap}
    \end{subfigure}}

    \caption{Large-scale distribution and co-occurrence of \sysname codes on 3{,}455 behavioral sentences extracted from 211 agent papers.}
    \label{fig:appendix:distribution_scale_combined}
\end{figure}

\section{Extension Protocol}
\label{sec:extension}
\sysname is meant to be a single shared codebook that grows as new agent designs appear, while making version changes explicit for comparison across papers. As reported in Section~\ref{sec:phase2}, the action and sub-action levels are relatively stable, with new additions occurring primarily at the leaf level. We support extension in two stages: the \extensiontool{} lets a downstream user grow their own copy of the codebook locally, and a public submission channel folds general-interest codes back into the shared release.

\paragraph{Local extension.}
When a user downloads \sysname, they receive the current released codebook as their starting point. As they analyze new trajectories, \extensiontool{} compares each observed behavior against the codebook: if an existing code fits, it assigns the code; if no existing code fits, it proposes a new leaf (or, more rarely, a new sub-action or top-level category) along with the position where the code should sit. The user accepts the suggestion to grow their local copy without blocking on the public codebook. We release \extensiontool{} as a Claude Code Skill in our open repository; any sufficiently capable LLM backend can serve as a drop-in replacement.

\paragraph{Public submission.}
If a user judges a locally added code to be of general interest, they submit it through a GitHub issue or pull request in the same repository. We review submissions and incorporate accepted ones into the next versioned release. 

\clearpage
\section{Automated Trace Analysis Tool}
\label{sec:appendix:skill}
\begin{figure}[!htbp]
    \centering
    \includegraphics[width=\textwidth]{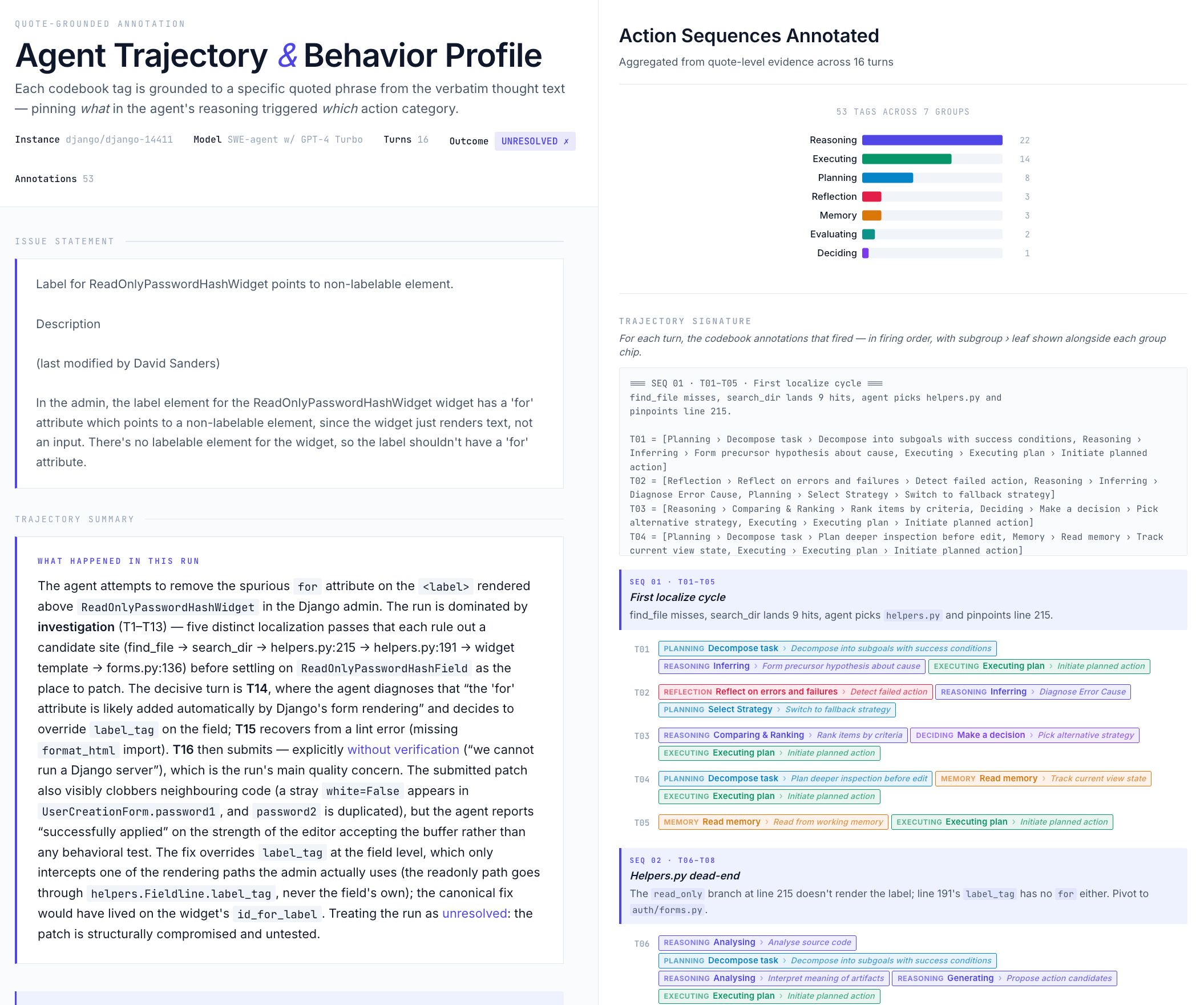}
    \caption{Interface of \autotraceqda{}.}
    \label{fig:appendix:interface}
\end{figure}

\autotraceqda{} (introduced in Section~\ref{sec:case-study}) takes a raw agent trajectory and, for each turn, assigns Action, Sub-action, and Leaf-category labels to its thought and action components, together with the verbatim quote behind every label. Figure~\ref{fig:appendix:interface} shows the rendered HTML report: the overall action distribution alongside a per-label quote panel. Refer to our website for more details.

\subsection{Raw Trajectory Sources and Examples}
\label{sec:raw-math-trajectories}

\paragraph{AG2/MathChat trajectories used in our analysis.}
The 100 AG2/MathChat trajectories analyzed in Section~\ref{sec:case-study-1} come from the experimental data released with MAST~\citep{cemri2025multiagentllmsystemsfail} and cover 77 distinct GSM-Plus math word problems. The upstream trajectory directory is publicly available.\footnote{MAST AG2 experimental trajectories: \url{https://github.com/multi-agent-systems-failure-taxonomy/MAST/tree/main/traces/AG2/experiments}.} The following examples link directly to the original JSON files, which contain the problem statements, agent trajectories, and evaluation metadata before our preprocessing and annotation. Problem descriptions below are abridged; answers are the reference answers recorded in the source files.

\begin{itemize}[leftmargin=*,nosep]
    \item \textbf{Income and spending.} Gerald earns \$30 per day, works for seven days, and saves \$110. How much did he spend? Reference answer: \$100. This GSM-Plus example reverses the operation in the seed question. \href{https://github.com/multi-agent-systems-failure-taxonomy/MAST/blob/main/traces/AG2/experiments/trajs_gpt-4o_impr_prompt_orig_topology_44/018efed1-9951-5512-a991-d2115e718547.json}{\underline{Original JSON trajectory}}.
    \item \textbf{Running speed with irrelevant information.} John runs 60 miles over three days, spending three hours on the first day and half as long on each remaining day. He rests for two hours after each run. What is his running speed? Reference answer: 10 miles per hour; the rest periods do not enter the calculation. \href{https://github.com/multi-agent-systems-failure-taxonomy/MAST/blob/main/traces/AG2/experiments/trajs_gpt-4_impr_prompt_impr_topology_43/09a5652f-b49c-5f8e-aa2f-4a3634d93164.json}{\underline{Original JSON trajectory}}.
\end{itemize}

\paragraph{HyperAgent trajectories used in our analysis.}
The 100 HyperAgent trajectories analyzed in Section~\ref{sec:case-study-1} were sampled from the MAST HyperAgent collection.\footnote{MAST HyperAgent trajectories: \url{https://github.com/multi-agent-systems-failure-taxonomy/MAST/tree/main/traces/HyperAgent}.} MAST reformatted these records from the HyperAgent submission to SWE-bench Lite dated September 25, 2024. The original trajectory assets are listed in the upstream submission metadata.\footnote{HyperAgent upstream trajectory location: \url{https://github.com/SWE-bench/experiments/blob/main/evaluation/lite/20240925_hyperagent_lite1/metadata.yaml}.} None of the 100 sampled instance identifiers appears in the upstream submission's list of resolved instances.\footnote{HyperAgent upstream evaluation results: \url{https://github.com/SWE-bench/experiments/blob/main/evaluation/lite/20240925_hyperagent_lite1/results/results.json}.} For a concrete example from our sample, \texttt{astropy\_\_astropy-14182} concerns support for header rows in reStructuredText table output; its \href{https://github.com/multi-agent-systems-failure-taxonomy/MAST/blob/main/traces/HyperAgent/astropy__astropy-14182.json}{\underline{MAST JSON trajectory}} includes the issue description, agent log lines, and task metadata.

\textbf{Concrete bug-repair example: missing SQL sorting clauses.} In \texttt{django\_\_django-11001}, Django incorrectly removes distinct multiline \texttt{RawSQL} sorting expressions when their final lines match. The duplicate-detection regular expression sees only the final line, causing the generated query to lose a requested sorting clause. The \href{https://github.com/multi-agent-systems-failure-taxonomy/MAST/blob/main/traces/HyperAgent/django__django-11001.json}{\underline{original trajectory}} records the following repair attempt:
\begin{itemize}[leftmargin=*,nosep]
    \item \textbf{Locate and propose a fix.} HyperAgent investigates \texttt{SQLCompiler.get\_order\_by()} in \nolinkurl{django/db/models/sql/compiler.py}. It proposes collapsing newlines and repeated whitespace before extracting the comparison key, so duplicate detection compares the full expression.
    \item \textbf{Check the proposed behavior.} After attempts to run the Django test fail because \texttt{manage.py} cannot be found, the agents create a standalone \texttt{MockSQLCompiler} test. The test checks that all three supplied sorting clauses are retained. The Executor reports that this test passes; verification in the recorded run is limited to this simulation.
\end{itemize}

\section{Additional Validation}
\label{sec:additional-validation}

\subsection{Agreement Across Models and Humans}
\label{sec:model-agreement}

\paragraph{Annotator models and calibration.} We applied the same codebook and annotation protocol with Claude Opus 5, GPT-5.6-sol through Codex, and DeepSeek-V4-Pro through OpenCode, with no communication between runs. 

\paragraph{Matched-span evaluation.} The three models annotated 10 additional trajectories, with 523 spans matched across all three models; 330 spans received the same action and sub-action from every model. Table~\ref{tab:model-agreement} reports pairwise Cohen's $\kappa$ and three-way Fleiss' $\kappa$ for labels on matched spans. 
To separate the annotator from the model generating the trace, we also annotated 10 public SWE-agent trajectories generated with GPT-4o (2024-05-13). 
Agreement remained high at the action level and decreased on finer levels.

\begin{table}[!htbp]
\centering\small
\caption{Label agreement across three annotators. C: Claude Opus 5; G: GPT-5.6-sol; D: DeepSeek-V4-Pro. Pairwise columns report Cohen's $\kappa$; the final column reports Fleiss' $\kappa$. The two trajectory samples are evaluated separately.}
\label{tab:model-agreement}
\begin{tabular}{llrrrr}
\toprule
Trajectory sample & Level & C--G & C--D & G--D & All three \\
\midrule
10 additional trajectories (generated by Claude model) & Action & .851 & .738 & .685 & .759 \\
 & Sub-action & .804 & .685 & .640 & .711 \\
 & Leaf & .755 & .624 & .588 & .653 \\
\midrule
10 GPT-4o-generated trajectories & Action & .853 & .824 & .857 & .843 \\
 & Sub-action & .771 & .729 & .817 & .762 \\
 & Leaf & .713 & .678 & .663 & .675 \\
\bottomrule
\end{tabular}
\end{table}

\begin{table}[!htbp]
\centering
\caption{Human--model agreement on 36 quotes: raw agreement / Cohen's $\kappa$. }
\label{tab:human-model}
\begin{tabular}{lrrr}
\toprule
Level & Claude Opus 5 & GPT-5.6-sol & DeepSeek-V4-Pro \\
\midrule
Action & 72.2\% / .660 & 83.3\% / .791 & 83.3\% / .790 \\
Sub-action & 61.1\% / .555 & 63.9\% / .588 & 69.4\% / .640 \\
Leaf & 30.6\% / .272 & 30.6\% / .281 & 36.1\% / .334 \\
\bottomrule
\end{tabular}
\end{table}

\begin{table}[!htbp]
\centering
\caption{Human-human inter-annotator agreement on trajectory annotations (90 quotes).}
\label{tab:human-human}
\begin{tabular}{lrrr}
\toprule
Level & Raw agreement & Cohen's $\kappa$ & Reported 95\% CI \\
\midrule
Action & .859 & .814 & [.70, .91] \\
Sub-action & .484 & .436 & [.32, .56] \\
Leaf & .297 & .268 & [.16, .38] \\
\bottomrule
\end{tabular}
\end{table}

\paragraph{Behavior extraction.} We also repeated the discovery step on the same five papers with the three models, obtaining 503 behavior instances in total. 
Agreement scores were $0.668$ for Claude--Codex, $0.704$ for Claude--DeepSeek, and $0.693$ for Codex--DeepSeek. 

\subsection{Reproducibility of Behavioral Profiles}
\label{sec:profile-validation}

\paragraph{Independent trajectory samples.} We compared profiles using Jensen--Shannon distance, which is zero for identical distributions. 
Across 2,000 resamples with disjoint sets of 25 trajectories per profile, within-agent median distances were $0.060$--$0.069$; the reported between-agent distances ranged from $0.149$ to $0.244$. 
The estimated probability that a between-agent distance exceeded a within-agent distance was $0.9997$. This supports stability within the sampled configurations and task pools. 

\begin{table}[!htbp]
\centering\small
\caption{Agreement in action rankings across annotator models on 25 trajectories per agent. Pairwise ranges cover the three model pairs.}
\label{tab:profile-agreement}
\begin{tabular}{lrrr}
\toprule
Corpus & Spearman $\rho$ & Kendall's $W$ & $p$ for $W$ \\
\midrule
SWE-agent & .976--.988 & .989 & .0016 \\
AG2 & .966--.988 & .970 & .0019 \\
HyperAgent & .879--.952 & .941 & .0026 \\
\bottomrule
\end{tabular}
\end{table}

\subsection{Exploratory Association with Task Outcomes}
\label{sec:pattern-validation}

We applied the pipeline to 32 public SWE-agent trajectories, comprising 16 resolved and 16 unresolved tasks and 2,289 tags. 
Reflection's share in the last three turns was higher in unresolved trajectories. Specifically, the ``recognize a knowledge boundary, then submit anyway'' pattern occurred in 4/16 unresolved trajectories and 1/16 resolved trajectories. 

\begin{table}[!htbp]
\centering\footnotesize
\setlength{\tabcolsep}{4pt}
\caption{Exploratory outcome comparison on 32 trajectories. R: resolved; U: unresolved. }
\label{tab:outcome-pilot}
\begin{tabularx}{\linewidth}{@{}>{\raggedright\arraybackslash}Xrrrr@{}}
\toprule
Measure & R & U & U$-$R [95\% CI] & AUC \\
\midrule
Reflection share, whole trace & .098 & .091 & $-.008$ [$-.041$, $.026$] & .484 \\
Reflection share, final three turns & .028 & .095 & $.067$ [$.012$, $.122$] & .713 \\
Gold check at any point & .625 & .438 & $-.188$ [$-.500$, $.188$] & --- \\
\bottomrule
\end{tabularx}
\end{table}

\clearpage

\section{Action Space Codebook}
\label{sec:full-codebook}

\begingroup
\raggedbottom
\let\codebookparagraph\paragraph
\renewcommand{\paragraph}{\Needspace{8\baselineskip}\codebookparagraph}
\let\codebooksubsection\subsection
\renewcommand{\subsection}{\Needspace{13\baselineskip}\codebooksubsection}
\newcolumntype{K}[1]{>{\raggedright\arraybackslash}p{#1}}
\newcommand{\codebookheader}{%
  \specialrule{1pt}{0pt}{0pt}
  \rowcolor{tabHead}\textcolor{white}{\textbf{Specialization}} & \textcolor{white}{\textbf{Example \& Quote}} \\
  \specialrule{1pt}{0pt}{0pt}
  \endfirsthead
  \specialrule{1pt}{0pt}{0pt}
  \rowcolor{tabHead}\textcolor{white}{\textbf{Specialization (cont.)}} & \textcolor{white}{\textbf{Example \& Quote}} \\
  \specialrule{1pt}{0pt}{0pt}
  \endhead
  \specialrule{1pt}{0pt}{0pt}
  \endlastfoot
}
\rowcolors{2}{tabAlt}{white}

\subsection{\cat{Grounding} Sub-actions}

\paragraph{Interact with users.}
\begin{longtable}{K{\dimexpr0.34\linewidth-2\tabcolsep\relax} K{\dimexpr0.66\linewidth-2\tabcolsep\relax}}
\codebookheader
\code{Accept instructions from humans} & Receive a task or command from a human user (e.g., ``book me a flight'' or ``translate this paragraph''); ``agent receives task queries and deliverables, completing tasks through multi-turn interactions'' (P16); ``demonstrate clear intent understanding, partial progress'' (P5). \\
\code{Ask for clarification from people} & Proactively ask the human when the request is unclear (e.g., ``Which file did you mean?''). \\
\code{Communicate task outcome in natural language} & Agent informs the user of the result, e.g., ``Buy a nice rich navy bathing dress'' (P3); WiFi-off notification example (P3). \\
\code{Communicate via visualization} & ``CellAgent generated differential expression and marker gene visualizations, enabling intuitive interpretation of cluster identities'' (P1). \\
\code{Communicate via structured format} & ``generate outputs in a JSON format (or lists of JSON objects)'' (P2). \\
\code{Communicate refusal/inability} & ``appropriately reject queries that exceed their knowledge boundaries or conflict with their role settings'' (P22); ``recognize and refuse queries that conflict with their role knowledge'' (P22); ``providing clear refusal responses with appropriate explanations'' (P22). \\
\code{Express tone to user} & ``models exhibit a strong default tendency to excessively affirm, apologize, and express gratitude'' (P11). \\
\code{Disclose self-information to user} & ``identity disclosure, S2S systems often proactively mention that they are intelligent assistants'' (P11). \\
\end{longtable}

\paragraph{Interact with physical environments.}
\begin{longtable}{K{\dimexpr0.34\linewidth-2\tabcolsep\relax} K{\dimexpr0.66\linewidth-2\tabcolsep\relax}}
\codebookheader
\code{Perceive physical environment} & Convert images/sensor/audio readings to text via VLMs so a text-based LLM can consume them; ``Explore skill enhances the LLM-based executor's ability to guide the agent in room exploration by sampling navigation goals from traversable areas'' (P33); ``semantic mapping module receives egocentric visual observations\ldots{} processed into a depth map and instance segmentation using a UNet and a MaskRCNN'' (P34). \\
\code{Affect physical environments} & Send language commands to a robot arm to move things in the real world; ``the action Play is utilized to interact with the environment or request re-planning of the current plan S'' (P34). \\
\end{longtable}

\paragraph{Interact with digital environments.}
\begin{longtable}{K{\dimexpr0.34\linewidth-2\tabcolsep\relax} K{\dimexpr0.66\linewidth-2\tabcolsep\relax}}
\codebookheader
\code{Navigate digital interfaces} & Browse websites, scroll pages, navigate menus, move within apps; ``navigate multiple websites, extract information from both structured and unstructured sources'' (P2). \\
\code{Modify digital objects} & Annotate UI components by adding \texttt{data-testid} attributes: ``\texttt{annotate\_interactive\_components(file, strategy=`add data-testid')}'' (P28). \\
\code{Issue operational commands} & Invoke a structured API or data endpoint: ``Game Control: \texttt{get\_game\_state}, \texttt{press\_buttons}, and \texttt{navigate\_to} for direct game control'' (P25). \\
\end{longtable}

\paragraph{Interact with other agents.}
\begin{longtable}{K{\dimexpr0.34\linewidth-2\tabcolsep\relax} K{\dimexpr0.66\linewidth-2\tabcolsep\relax}}
\codebookheader
\code{Monitor peer agent's state or decision} & Observe another agent's chosen action before making one's own decision---a prerequisite for oversight, override, or trust-based deference in multi-agent systems: ``observes both $S_0$ and $A_0$, and decides whether to override the AI's treatment or not'' (P33). \\
\code{Receive feedback from peer agent} & Evaluator provides natural-language feedback (P1). \\
\code{Send message to peer agent} & ``agent engages in multi-turn interactions\ldots{} to generate deliverables'' (P16). \\
\code{Recommend action to peer agent} & Communicate a proposed action to a peer agent for their consideration, knowing the peer may accept or override it---agent-to-agent advisory communication: ``the AI recommends one of eight possible treatments, which is then reviewed and potentially overridden by the clinician'' (P33). \\
\code{Override peer agent's decision} & Replace another agent's selected action with one's own decision---the active side of human-in-the-loop oversight or hierarchical agent control: ``If the clinician overrides the AI, the patient outcome $Y$ is determined by $S_0$ and the alternative treatment $H_0$ suggested by the clinician'' (P33). \\
\code{Dispatch task to sub-agent} & ``A central orchestrator maintains a high-level route plan while dynamically dispatching sub-agents based on game context'' (P25). \\
\code{Argue or debate with peer agent} & Multi-agent argumentation toward a better answer; have multiple agents argue different sides to reach a better answer. \\
\end{longtable}

\paragraph{Augment with external computation.}
\begin{longtable}{K{\dimexpr0.34\linewidth-2\tabcolsep\relax} K{\dimexpr0.66\linewidth-2\tabcolsep\relax}}
\codebookheader
\code{Execute code} & Test Runner detects logic inconsistencies (P28); ``Test Runner executes the script to detect logical inconsistencies or failed assertions'' (P28). \\
\code{Invoke specialized computation tool} & Use calculator/heavy-compute tools to handle math beyond the LLM; \texttt{run\_shell\_command} tool (425 and 362 invocations, respectively) (P16); ``necessitates over 90 precise tool calls, significantly raising the bar'' (P16); ``deliverables are synced to a Docker-based remote sandbox\ldots{} emulates human computer operations'' (P16). \\
\code{Invoke visual inspection tool} & Zoom-in tool for region-of-interest (P4): ``proactively invokes the Zoom-in tool for a targeted examination of the specific region of interest'' (P4). \\
\end{longtable}

\subsection{\cat{Planning} Sub-actions}

\paragraph{Decompose task.}
\begin{longtable}{K{\dimexpr0.34\linewidth-2\tabcolsep\relax} K{\dimexpr0.66\linewidth-2\tabcolsep\relax}}
\codebookheader
\code{Decompose into subtasks} & ``LLM-based planner to decompose the specified language instruction $L$ into a sequence of subtasks $S=[S_0, S_1, \ldots, S_n]$'' (P34); ``Decomposing hard problems into subproblems often makes them easier and more efficient to solve'' (P17); ``Decomposes complex user requests into manageable subtasks, an Executor that carries out the analysis by generating and running code, and an Evaluator that assesses the quality of the results'' (P1); ``formulate plans, decompose problems into sub-steps'' (P2); \texttt{todo\_write} tool (15 invocations by GPT-5.2) (P16). \\
\code{Decompose into subgoals with success conditions} & ``LLM decomposes the task into a sequence of subgoals, each paired with an executable success-condition function'' (P25). \\
\code{Decompose by role specialization} & ``splitting the instruction-following challenge into distinct reasoning and grounding roles handled by a reasoner agent and an actor agent'' (P34); ``Specialists each solve a distinct task type, receiving only the relevant portion of the input'' (P17). \\
\end{longtable}

\paragraph{Formulate a workflow or plan.}
\begin{longtable}{K{\dimexpr0.34\linewidth-2\tabcolsep\relax} K{\dimexpr0.66\linewidth-2\tabcolsep\relax}}
\codebookheader
\code{Formulate a high-level plan} & ``METRO starts with the identification of the strategic action $a_i$ for every expert utterance $u_i$ in transcript $D$'' (P10); ``summarizing it into a high-level planning directive that emphasizes cumulative temporal effects'' (P8). \\
\code{Formulate an analysis workflow} & ``the Planner accurately interprets user intent and formulates a comprehensive analysis workflow'' (P1). \\
\code{Plan navigation through environment} & ``planning to navigate the web; tasks require navigation through an average of 4.2 web pages'' (P2). \\
\code{Plan function or tool use} & ``I can use the pandas method \texttt{corr()} to calculate the Pearson correlation coefficient'' (P32). \\
\code{Plan code or artifact structure} & ``Planning: HTML Structure, CSS Styling, JavaScript Functionality'' (P28). \\
\code{Formulate plan from template} & ``HLSTuner formulates a detailed plan that specifies: (1) the combination of HLS directives, (2) target code segments, and (3) the insertion actions'' (P29). \\
\end{longtable}

\paragraph{Select strategy.}
\begin{longtable}{K{\dimexpr0.34\linewidth-2\tabcolsep\relax} K{\dimexpr0.66\linewidth-2\tabcolsep\relax}}
\codebookheader
\code{Select among candidate strategies} & ``selects effective HLS directive combination strategies and inserts directives within the specific structure (e.g., loops and arrays)'' (P29). \\
\code{Switch to fallback strategy} & ``incorporating a dummy score prediction as a fallback mechanism'' (P11). \\
\end{longtable}

\paragraph{Modify plan.}
\begin{longtable}{K{\dimexpr0.34\linewidth-2\tabcolsep\relax} K{\dimexpr0.66\linewidth-2\tabcolsep\relax}}
\codebookheader
\code{Replan dynamically based on feedback} & ``RequireReplan provides the LLM-based executor with the capability to dynamically adjust the plan'' (P34). \\
\code{Refine requirements} & ``Refine requirement based on validated test cases to ensure alignment'' (P28). \\
\end{longtable}

\subsection{\cat{Reasoning} Sub-actions}

\paragraph{Generating.}
\begin{longtable}{K{\dimexpr0.34\linewidth-2\tabcolsep\relax} K{\dimexpr0.66\linewidth-2\tabcolsep\relax}}
\codebookheader
\code{Generate candidate options} & Brainstorm one or more possible next moves; ``Running multiple candidate pipelines for each analysis step'' (P1); ``generates solutions based on analogies to unrelated projects, which resemble few-shot prompting'' (P28); ``Fork the current generative state and generate a new parallel trajectory from this point of high uncertainty'' (P4); ``the LLM reinterprets the actions from breadth logic\ldots{} summarizing them into a concise next-step strategy'' (P8). \\
\code{Generate structured artifacts} & ``Each agent then emits a set of initial codes; cluster semantically similar codes and generate preliminary themes'' (P27); ``automatically generates candidate user-facing requirements'' (P28); ``Create JavaScript functionality handling empty display and consecutive operator clicks'' (P28); ``break down the given character description into multiple atomic pieces of knowledge'' (P22). \\
\code{Generate evaluations} & ``LLM Group for multifaceted evaluation generates diverse debugging instructions'' (P29); ``For every binary score $z^{(q)}_{i,j}$ from the judge, there is a corresponding explanation $e^{(q)}_{i,j}$'' (P3). \\
\end{longtable}

\paragraph{Analysing.}
\begin{longtable}{K{\dimexpr0.34\linewidth-2\tabcolsep\relax} K{\dimexpr0.66\linewidth-2\tabcolsep\relax}}
\codebookheader
\code{Analyse artifact structure and behavior} & ``Analyze frontend framework and interactive components'' (P28); ``analyzes the project's core functionalities and their interactions with UI elements by reading the source code'' (P28). \\
\code{Detect patterns or trends in data} & ``Identifying patterns, directions, or changes in data over time or across contexts'' (P2); ``Measuring relationships or associations between two or more variables'' (P2). \\
\code{Interpret meaning of artifacts} & ``logical reasoning to interpret papers and code; mathematical reasoning to modify and run code; causal reasoning to infer scientific insights from results'' (P11). \\
\code{Classify inputs into categories} & ``delegator can determine which of the $k$ tasks the input string belongs to by observing only a constant number of metadata tokens'' (P17). \\
\end{longtable}

\paragraph{Explaining.}
\begin{longtable}{K{\dimexpr0.34\linewidth-2\tabcolsep\relax} K{\dimexpr0.66\linewidth-2\tabcolsep\relax}}
\codebookheader
\code{Explain reasoning or outcomes} & ``Explain the failure from a user requirement perspective'' (P16). \\
\end{longtable}

\paragraph{Summarizing/Distilling.}
\begin{longtable}{K{\dimexpr0.34\linewidth-2\tabcolsep\relax} K{\dimexpr0.66\linewidth-2\tabcolsep\relax}}
\codebookheader
\code{Summarize recent observations and trajectories} & Condense recent observations and trajectories into key takeaways. \\
\end{longtable}

\paragraph{Inferring.}
\begin{longtable}{K{\dimexpr0.34\linewidth-2\tabcolsep\relax} K{\dimexpr0.66\linewidth-2\tabcolsep\relax}}
\codebookheader
\code{Infer hidden state from observable evidence} & ``intelligently infers hidden information through game mechanics: damage calculations reveal stat distributions, move priority ordering constrains speed ranges'' (P25). \\
\code{Infer causal relationship} & ``determine the most plausible event or factor accounting for this variability'' (P2). \\
\code{Infer structure from indirect evidence} & ``agents must infer the directory layout by inspecting package structures and README files'' (P11). \\
\code{Infer errors} & ``$M_q$ has the capacity to identify exactly one bug at a time'' (P17); ``The analysis LLM then examines the error causes and provides debugging instructions'' (P29). \\
\end{longtable}

\paragraph{Comparing \& Ranking.}
\begin{longtable}{K{\dimexpr0.34\linewidth-2\tabcolsep\relax} K{\dimexpr0.66\linewidth-2\tabcolsep\relax}}
\codebookheader
\code{Compare values across sources} & ``Quantifying occurrences and comparing values across sources or categories'' (P2). \\
\code{Rank items by criteria} & ``Ordering items or facts based on specific criteria or importance'' (P2). \\
\end{longtable}

\paragraph{Contextualizing.}
\begin{longtable}{K{\dimexpr0.34\linewidth-2\tabcolsep\relax} K{\dimexpr0.66\linewidth-2\tabcolsep\relax}}
\codebookheader
\code{Package prior reasoning as context for subsequent calls} & Feed the reasoning so far as context to the next LLM call. \\
\code{Construct structured context object} & ``global interactive multi-behavior overlapping network is constructed based on the simulated behaviors of reposting and reposting with a comment, as well as following'' (P21); ``construct prompt templates for creating reader agents based on user attributes $u_a$ and user historical posts'' (P21). \\
\code{Configure agent persona or role-conditioning} & ``I want you to play as \{role\}, imitating \{role\}'s personality and values'' (P22); ``maintain its role-playing ability even when refusing to answer'' (P22); ``A pool of $k=4$ role-conditioned GPT-4o agents are each given the full interview transcript as input'' (P27); ``others may retain role-specific perspectives to support diverse theme formulation'' (P27). \\
\code{Assign roles in a multi-agent team} & ``specifying a scope, or role, for an agent allows it to be treated as a computational object\ldots{} map each one to an employee in a human organization'' (P17); ``assembling a team of three LLM roles---analyst, coder, and tester---responsible for analysis, coding, and testing'' (P28); ``assigns diverse roles to multiple agents to efficiently decompose complex tasks'' (P28). \\
\end{longtable}

\paragraph{Combining \& Synthesis.}
\begin{longtable}{K{\dimexpr0.34\linewidth-2\tabcolsep\relax} K{\dimexpr0.66\linewidth-2\tabcolsep\relax}}
\codebookheader
\code{Combine information from multiple sources} & ``combine information from multiple sources\ldots{} to produce a coherent solution'' (P2); ``the final results from all subtasks are synthesized to meet the user's requirements'' (P1); ``Determining a representative value that summarises numerical data collected from multiple sources'' (P2). \\
\code{Aggregate observations into a structured representation} & ``supplementary semantic map $M'_t$, which aggregates the information from $M_t$ over successive time steps. The intuition resembles a form of majority voting'' (P34); ``aggregates prediction results from multiple tools to generate the final cell type labels'' (P1). \\
\end{longtable}

\paragraph{Filtering.}
\begin{longtable}{K{\dimexpr0.34\linewidth-2\tabcolsep\relax} K{\dimexpr0.66\linewidth-2\tabcolsep\relax}}
\codebookheader
\code{Filter information by threshold} & ``Selecting relevant information based on criteria, quality, or thresholds'' (P2). \\
\end{longtable}

\subsection{\cat{Retrieval} Sub-actions}

\paragraph{Retrieve from skill library.}
\begin{longtable}{K{\dimexpr0.34\linewidth-2\tabcolsep\relax} K{\dimexpr0.66\linewidth-2\tabcolsep\relax}}
\codebookheader
\code{Retrieve from skill library} & Grab a pre-built skill snippet (e.g., Minecraft ``chop tree'') from a library of ready-made code; find and load a ready-made code snippet from a pre-built library. \\
\end{longtable}

\paragraph{Retrieve from local corpus.}
\begin{longtable}{K{\dimexpr0.34\linewidth-2\tabcolsep\relax} K{\dimexpr0.66\linewidth-2\tabcolsep\relax}}
\codebookheader
\code{Retrieve from local corpus} & ``diverse filetype reading; read between 1 to 15 documents and/or tables'' (P2); ``read the README file---Phase 1: \nolinkurl{read_file('reproduction_package/readme.txt')}'' (P11). \\
\end{longtable}

\paragraph{Retrieve from external knowledge base.}
\begin{longtable}{K{\dimexpr0.34\linewidth-2\tabcolsep\relax} K{\dimexpr0.66\linewidth-2\tabcolsep\relax}}
\codebookheader
\code{Retrieve from external knowledge base} & ``LLM which generates buggy code by integrating retrieved error slices from BugRAG as context'' (P29); ``queries BugRAG to check for existing entries'' (P29); ``Recall three (03) relevant and distinct problems (different from the user task)'' (P28). \\
\end{longtable}

\paragraph{Retrieve from open web.}
\begin{longtable}{K{\dimexpr0.34\linewidth-2\tabcolsep\relax} K{\dimexpr0.66\linewidth-2\tabcolsep\relax}}
\codebookheader
\code{Retrieve from open web} & ``high reliance on web search to offload knowledge retrieval to external sources'' (P16). \\
\end{longtable}

\paragraph{Retrieve relevant context.}
\begin{longtable}{K{\dimexpr0.34\linewidth-2\tabcolsep\relax} K{\dimexpr0.66\linewidth-2\tabcolsep\relax}}
\codebookheader
\code{Retrieve relevant context} & ``LLM leverages retrieved HLS-related context to transform input C algorithms or natural language descriptions'' (P29). \\
\end{longtable}

\subsection{\cat{Memory} Sub-actions}

\paragraph{Store information.}
\begin{longtable}{K{\dimexpr0.34\linewidth-2\tabcolsep\relax} K{\dimexpr0.66\linewidth-2\tabcolsep\relax}}
\codebookheader
\code{Store information in working memory} & A scratchpad / quick-access buffer that holds recent inputs and intermediate results. \\
\code{Store episodic trajectories} & Record full action sequences (start to finish) for later training or review. \\
\code{Store knowledge in semantic memory} & Save general world facts not tied to any specific event. \\
\code{Store experiences in episodic memory} & Save past events as personal-diary-like episodes (e.g., which game was won, which plan failed). \\
\code{Store information in long-term memory} & Persist state and key information in external storage; ``Gemini-3-Pro\ldots{} \texttt{initialize\_memory\_bank} (7 times); Gemini-3-Pro\ldots{} \texttt{update\_memory\_bank} (22 times)\ldots{} attempts to persist state and key information externally'' (P16). \\
\code{Maintain curriculum library} & Keep a syllabus of mastered and upcoming skills, arranged by difficulty, so the agent learns in a sensible order. \\
\end{longtable}

\paragraph{Update information.}
\begin{longtable}{K{\dimexpr0.34\linewidth-2\tabcolsep\relax} K{\dimexpr0.66\linewidth-2\tabcolsep\relax}}
\codebookheader
\code{Update memory} & Update experiences---refinement of memories (represented in text) as the system encounters new scenarios. \\
\end{longtable}

\paragraph{Discard information.}
\begin{longtable}{K{\dimexpr0.34\linewidth-2\tabcolsep\relax} K{\dimexpr0.66\linewidth-2\tabcolsep\relax}}
\codebookheader
\code{Discard information from working memory} & ``the local memory is discarded upon successful completion of the subtask'' (P1). \\
\code{Discard redundant game state, keep summaries} & ``automatic context compaction to manage thousands of reasoning steps, preserving only LLM responses and action summaries while discarding redundant game state'' (P25). \\
\end{longtable}

\paragraph{Consolidate memory.}
\begin{longtable}{K{\dimexpr0.34\linewidth-2\tabcolsep\relax} K{\dimexpr0.66\linewidth-2\tabcolsep\relax}}
\codebookheader
\code{Consolidate working memory into long-term memory} & Discard-after-use strategy that retains only the clear and successful analysis path in global memory, preventing interference from intermediate trial-and-error history (P1). \\
\code{Compact context window} & ``automatic context compaction to manage the thousands of reasoning steps required, preserving only LLM responses and action summaries while discarding redundant game state'' (P25). \\
\end{longtable}

\paragraph{Read memory.}
\begin{longtable}{K{\dimexpr0.34\linewidth-2\tabcolsep\relax} K{\dimexpr0.66\linewidth-2\tabcolsep\relax}}
\codebookheader
\code{Read from working memory} & Check the scratchpad: grab stored intermediate results and state; ``IMPLICITMEMBENCH reframes evaluation from `what agents recall' to `what they automatically enact'\thinspace'' (P12). \\
\code{Read from long-term memory} & Importance-weighted retrieval of discoveries (P25): ``Persistent memory system storing discoveries (locations, NPCs, items, strategies) with importance-weighted retrieval'' (P25). \\
\end{longtable}

\subsection{\cat{Evaluating} Sub-actions}

\paragraph{Evaluating with gold.}
\begin{longtable}{K{\dimexpr0.34\linewidth-2\tabcolsep\relax} K{\dimexpr0.66\linewidth-2\tabcolsep\relax}}
\codebookheader
\code{Compare against gold reference} & ``we prompt LLM to review the correct HLS-C code, pairing buggy code segments with corresponding error messages to construct debugging CoT'' (P29); ``HLSFixer retests the corrected HLS design against the golden results to ensure semantic equivalence'' (P29). \\
\code{Score on gold criteria} & ``The Gold Checker evaluates each annotation on three binary criteria: equivalence, completeness, and correctness, where a score of 1 indicates the criterion is met'' (P28). \\
\end{longtable}

\paragraph{Evaluating with goals/requirements/constraints.}
\begin{longtable}{K{\dimexpr0.34\linewidth-2\tabcolsep\relax} K{\dimexpr0.66\linewidth-2\tabcolsep\relax}}
\codebookheader
\code{Goal-completion check} & ``Independently checks whether objectives are truly complete, preventing the orchestrator from advancing when the main agent incorrectly believes a task is finished'' (P25). \\
\code{Requirement-satisfaction check} & ``the final results from all subtasks are synthesized to meet the user's requirements'' (P1). \\
\code{Constraint / budget check} & ``HLSTuner enables QoR-aware reasoning to align optimization goals with hardware constraints'' (P29); ``if hardware utilization exceeds the budget'' (P29). \\
\code{Domain-rule / best-practice check} & ``codified best-practices, such as the standard order of operations (e.g., quality control must precede normalization)'' (P1); ``Agents share project context to ensure: Cross-file consistency; Non-conflicting test-ids'' (P28); ``Detect logical errors, missing functionalities, or violations of best practices'' (P16). \\
\end{longtable}

\paragraph{Evaluating without ground truth.}
\begin{longtable}{K{\dimexpr0.34\linewidth-2\tabcolsep\relax} K{\dimexpr0.66\linewidth-2\tabcolsep\relax}}
\codebookheader
\code{Score on quality dimensions} & ``we use GPT-4o as the default evaluator, each dimension scored on a scale of 0 to 2'' (P22); ``map the agent's deliverables to a score ranging from 0 to 10'' (P16); ``Evaluation Scores: Obtain $s^{(t)}=(C,D,T)$ on the trustworthiness dimensions'' (P27); ``the scoring agent (functioning as the judge) based on clarity, logical soundness, alignment with error messages, this agent selects the optimal suggestion'' (P29). \\
\code{Check rubric compliance} & ``Verify the implementation of every item listed in \{Rubrics\}'' (P16); ``list exactly which rubrics were not met'' (P16). \\
\code{Evaluate visual/behavioral correctness} & ``Evaluates dynamic behavior and visual correctness based on screenshots'' (P16); ``Analyze \{Visual Assets\} for UI elements, layout consistency, and text rendering'' (P16). \\
\code{Evaluate internal consistency} & ``task an evaluator agent with determining whether each theme is consistent with its supporting quotes'' (P27); ``In Phase 2, they inspect the provided code for potential inconsistencies; generating a reproducibility score on a scale from 1 to 4'' (P11). \\
\code{Evaluate intermediate results} & ``agents with domain expertise verify intermediate results and reduce errors'' (P28). \\
\code{Provide qualitative judgement} & ``Upon successful generation of a result, the Evaluator agent, ALLMe assesses the outcome and provides natural language feedback if it deems revisions are necessary'' (P1). \\
\code{Simulate counterfactual outcomes} & Mentally simulate an alternative scenario---either before acting (``if I do X, what happens?'') or post-hoc (``what would have happened if action X had been different?'')---including counterfactual trajectory generation for causal analysis: ``analyzing this effect necessitates counterfactual reasoning across three distinct scenarios'' (P33). \\
\code{Validate predicted issues} & ``The agent assesses the contextual applicability of potential bugs, reducing the probability that the LLM forcibly generates trivial results'' (P29). \\
\end{longtable}

\subsection{\cat{Deciding} Sub-actions}

\paragraph{Make a decision.}
\begin{longtable}{K{\dimexpr0.34\linewidth-2\tabcolsep\relax} K{\dimexpr0.66\linewidth-2\tabcolsep\relax}}
\codebookheader
\code{Make a decision according to memory} & ``The agent conditions its decisions on $I$, the current observation $s_t$, and a history buffer of $n$ past state--action pairs'' (P5). \\
\end{longtable}

\paragraph{Pick scores.}
\begin{longtable}{K{\dimexpr0.34\linewidth-2\tabcolsep\relax} K{\dimexpr0.66\linewidth-2\tabcolsep\relax}}
\codebookheader
\code{Select action by score} & argmax = highest score; softmax = probabilistic sample; majority vote across evaluators. \\
\end{longtable}

\paragraph{Decide accept or not.}
\begin{longtable}{K{\dimexpr0.34\linewidth-2\tabcolsep\relax} K{\dimexpr0.66\linewidth-2\tabcolsep\relax}}
\codebookheader
\code{Decline out-of-scope queries} & ``appropriately reject queries that exceed their knowledge boundaries or conflict with their role settings'' (P22). \\
\end{longtable}

\paragraph{Decide under uncertainty.}
\begin{longtable}{K{\dimexpr0.34\linewidth-2\tabcolsep\relax} K{\dimexpr0.66\linewidth-2\tabcolsep\relax}}
\codebookheader
\code{Fork trajectory at uncertainty} & ``Fork the current generative state and generate a new parallel trajectory from this point of high uncertainty'' (P4). \\
\end{longtable}

\subsection{\cat{Executing} Sub-actions}

\paragraph{Executing plan.}
\begin{longtable}{K{\dimexpr0.34\linewidth-2\tabcolsep\relax} K{\dimexpr0.66\linewidth-2\tabcolsep\relax}}
\codebookheader
\code{Execute strategy} & ``an insertion agent executes this plan for HLS-C optimization'' (P29). \\
\end{longtable}

\paragraph{Executing debug.}
\begin{longtable}{K{\dimexpr0.34\linewidth-2\tabcolsep\relax} K{\dimexpr0.66\linewidth-2\tabcolsep\relax}}
\codebookheader
\code{Adopt debugging instructions} & ``Operating under strict instruction adherence, this agent adopts the instructions to implement debugging'' (P29). \\
\code{Rewrite code after bug fix} & ``debugging specialist rewrites the entire code every time it fixes a bug'' (P17). \\
\end{longtable}

\paragraph{Terminating.}
\begin{longtable}{K{\dimexpr0.34\linewidth-2\tabcolsep\relax} K{\dimexpr0.66\linewidth-2\tabcolsep\relax}}
\codebookheader
\code{Provide final answer} & ``Final Answer: The average number of lines per scene in the \texttt{ep\_7.csv} dataset is approximately 6.02'' (P32); ``produces a final answer and terminates the rollout; enclose it within \texttt{<answer></answer>} tags'' (P4). \\
\code{Generate refusal} & ``providing clear refusal responses with appropriate explanations'' (P22). \\
\end{longtable}

\subsection{\cat{Reflecting} Sub-actions}

\paragraph{Reflect on errors and failures.}
\begin{longtable}{K{\dimexpr0.34\linewidth-2\tabcolsep\relax} K{\dimexpr0.66\linewidth-2\tabcolsep\relax}}
\codebookheader
\code{Diagnose failure against ground truth} & ``Analyzes stuck states by comparing current situation against ground truth sources (porymap data, knowledge base) to diagnose navigation failures'' (P25). \\
\code{Inspect error pattern} & ``an inspection agent examines the erroneous code and error messages parsed from the HLS tool test results'' (P29). \\
\code{Analyze log to formulate fix instructions} & Reasoning-to-instruction on HLS log (P29): ``an analysis agent adopts a reasoning-to-instruction method, analyzing the HLS log to formulate error modification actions'' (P29); ``This model formulates explicit modification instructions with detailed analysis'' (P29); ``Provide explicit instructions on what must be rectified in the next iteration'' (P16). \\
\code{Reflect on failed episodes for prior knowledge} & ``RequireReplan\ldots{} dynamically adjust the plan, improving the robustness to exceptions'' (P26); ``the grounded prior knowledge prevents the agents from repetitive errors and facilitates grounded exception handling'' (P34); ``This allows the agent to learn from mistakes in realtime and avoid repeating errors before the local memory is discarded'' (P1). \\
\code{Self-correct step implementation} & ``Modify step implementation based on error message'' (P28); ``In case of an execution error $E(c_i)$, the Executor autonomously performs self-correction to produce a valid code version'' (P1); ``When the initial attempt fails\ldots{} HLSTuner activates an iterative refinement incorporating current directives and the resulting QoR'' (P29). \\
\end{longtable}

\paragraph{Reflect on self-outcomes.}
\begin{longtable}{K{\dimexpr0.34\linewidth-2\tabcolsep\relax} K{\dimexpr0.66\linewidth-2\tabcolsep\relax}}
\codebookheader
\code{Self-reflect} & ``The results of the execution are fed back, enabling GPT-4 to refine its responses and propose further iterations'' (P32); ``This iterative cycle continues until GPT-4 determines that the accumulated information suffices to conclusively answer the problem'' (P32); ``This self-reflective optimization loop iterates to enhance precision, and the final results from all subtasks are synthesized'' (P1); ``the optimization process runs three iterations, each invoking a different algorithm'' (P1). \\
\code{Reflect on proposed fix} & ``reflects on the code after assuming the fix to ensure the modification is reasonable'' (P29). \\
\code{Pre-action self-check} & ``These steps are labeled as reflective, indicating the model is engaging in self-monitoring or explicit reasoning before issuing commands'' (P1). \\
\code{Refine strategy iteratively} & Iterative refinement using current QoR (P29): ``When the initial attempt fails\ldots{} HLSTuner activates an iterative refinement incorporating current directives and the resulting QoR'' (P29). \\
\end{longtable}

\paragraph{Reflect on external feedback.}
\begin{longtable}{K{\dimexpr0.34\linewidth-2\tabcolsep\relax} K{\dimexpr0.66\linewidth-2\tabcolsep\relax}}
\codebookheader
\code{Receive and integrate external feedback} & ``the Evaluator agent ALLMe assesses the outcome and provides natural language feedback if it deems revisions are necessary'' (P1); ``feedback from the feedback agent is used to iteratively refine and improve the generated themes'' (P27); ``leverage feedback to markedly improve performance\ldots{} feedback-driven self-correction'' (P16); ``The Evaluator drives a self-reflective optimization mechanism, which leverages automated evaluation methods for various analysis tasks to iteratively refine outcomes, replacing subjective manual assessments'' (P1). \\
\end{longtable}

\subsection{\cat{Learning} Sub-actions}

\paragraph{Learning reasoning.}
\begin{longtable}{K{\dimexpr0.34\linewidth-2\tabcolsep\relax} K{\dimexpr0.66\linewidth-2\tabcolsep\relax}}
\codebookheader
\code{Update reasoning via prompt update} & Rewrite the agent's own prompt template; learn a better way to reason via prompt rewrite. \\
\end{longtable}

\paragraph{Learning grounding.}
\begin{longtable}{K{\dimexpr0.34\linewidth-2\tabcolsep\relax} K{\dimexpr0.66\linewidth-2\tabcolsep\relax}}
\codebookheader
\code{Update grounding via code-based skills} & Improve web-navigation code snippets; write or improve code that interacts with the outside world. \\
\code{Update retrieval procedures} & Improve how the agent searches for information (better keyword strategies, smarter ranking). \\
\end{longtable}

\paragraph{Learning knowledge.}
\begin{longtable}{K{\dimexpr0.34\linewidth-2\tabcolsep\relax} K{\dimexpr0.66\linewidth-2\tabcolsep\relax}}
\codebookheader
\code{Update source code as procedural memory} & Self-patch the agent's own source code to change its behavior. \\
\code{Update semantic memory with knowledge} & Expand error repository with new mnemonic (P29): ``the inspection agent identifies a new error type and integrates the slice into the error repository with a new mnemonic identifier'' (P29). \\
\code{Update memory from new experiences} & ``refinement of memories (represented in text) as the system encounters new scenarios'' (P17). \\
\code{Update hypothesis with new evidence} & ``allowing the model to update its textual understanding, refine its hypothesis, or even trigger further visual exploration'' (P4). \\
\end{longtable}

\paragraph{Learning LLM parameters.}
\begin{longtable}{K{\dimexpr0.34\linewidth-2\tabcolsep\relax} K{\dimexpr0.66\linewidth-2\tabcolsep\relax}}
\codebookheader
\code{Update parametric policy} & Change the model's internal weights through training; real learning by changing internal weights. \\
\code{Update LLM parameters via SL/RL/RLHF} & Use supervised, reinforcement, or human-feedback learning to adjust model weights. \\
\code{Update action parameters based on feedback} & Auto-adjust parameters from exception info (P1): ``automatically adjusts parameters using exception information to generate executable code'' (P1). \\
\end{longtable}

\paragraph{Learning instructions.}
\begin{longtable}{K{\dimexpr0.34\linewidth-2\tabcolsep\relax} K{\dimexpr0.66\linewidth-2\tabcolsep\relax}}
\codebookheader
\code{Infer instructions from input--output examples} & Extract the underlying rule from input/output examples for future use. \\
\end{longtable}

\endgroup

\clearpage
\section*{NeurIPS Paper Checklist}

\begin{enumerate}

\item {\bf Claims}
    \item[] Question: Do the main claims made in the abstract and introduction accurately reflect the paper's contributions and scope?
    \item[] Answer: \answerYes{}
    \item[] Justification: The abstract and introduction describe \sysname as a taxonomy with an open repository, constructed from 565 behavioral descriptions in 20 papers and demonstrated through two case studies (Sections~\ref{sec:method} and~\ref{sec:case-study}). The scope of the corpus and the limits of generalization are discussed in Appendix~\ref{sec:limitations}.
    \item[] Guidelines:
    \begin{itemize}
        \item The answer \answerNA{} means that the abstract and introduction do not include the claims made in the paper.
        \item The abstract and/or introduction should clearly state the claims made, including the contributions made in the paper and important assumptions and limitations. A \answerNo{} or \answerNA{} answer to this question will not be perceived well by the reviewers. 
        \item The claims made should match theoretical and experimental results, and reflect how much the results can be expected to generalize to other settings. 
        \item It is fine to include aspirational goals as motivation as long as it is clear that these goals are not attained by the paper. 
    \end{itemize}

\item {\bf Limitations}
    \item[] Question: Does the paper discuss the limitations of the work performed by the authors?
    \item[] Answer: \answerYes{}
    \item[] Justification: Appendix~\ref{sec:limitations} discusses the scope of the construction corpus, differences between paper descriptions and execution traces, and potential bias from LLM-assisted construction. Appendix~\ref{sec:impacts} also discusses limits of behavioral profiles and reproducibility when commercial models change.
    \item[] Guidelines:
    \begin{itemize}
        \item The answer \answerNA{} means that the paper has no limitation while the answer \answerNo{} means that the paper has limitations, but those are not discussed in the paper. 
        \item The authors are encouraged to create a separate ``Limitations'' section in their paper.
        \item The paper should point out any strong assumptions and how robust the results are to violations of these assumptions (e.g., independence assumptions, noiseless settings, model well-specification, asymptotic approximations only holding locally). The authors should reflect on how these assumptions might be violated in practice and what the implications would be.
        \item The authors should reflect on the scope of the claims made, e.g., if the approach was only tested on a few datasets or with a few runs. In general, empirical results often depend on implicit assumptions, which should be articulated.
        \item The authors should reflect on the factors that influence the performance of the approach. For example, a facial recognition algorithm may perform poorly when image resolution is low or images are taken in low lighting. Or a speech-to-text system might not be used reliably to provide closed captions for online lectures because it fails to handle technical jargon.
        \item The authors should discuss the computational efficiency of the proposed algorithms and how they scale with dataset size.
        \item If applicable, the authors should discuss possible limitations of their approach to address problems of privacy and fairness.
        \item While the authors might fear that complete honesty about limitations might be used by reviewers as grounds for rejection, a worse outcome might be that reviewers discover limitations that aren't acknowledged in the paper. The authors should use their best judgment and recognize that individual actions in favor of transparency play an important role in developing norms that preserve the integrity of the community. Reviewers will be specifically instructed to not penalize honesty concerning limitations.
    \end{itemize}

\item {\bf Theory assumptions and proofs}
    \item[] Question: For each theoretical result, does the paper provide the full set of assumptions and a complete (and correct) proof?
    \item[] Answer: \answerNA{}
    \item[] Justification: The paper does not present formal theoretical results.
    \item[] Guidelines:
    \begin{itemize}
        \item The answer \answerNA{} means that the paper does not include theoretical results. 
        \item All the theorems, formulas, and proofs in the paper should be numbered and cross-referenced.
        \item All assumptions should be clearly stated or referenced in the statement of any theorems.
        \item The proofs can either appear in the main paper or the supplemental material, but if they appear in the supplemental material, the authors are encouraged to provide a short proof sketch to provide intuition. 
        \item Inversely, any informal proof provided in the core of the paper should be complemented by formal proofs provided in appendix or supplemental material.
        \item Theorems and Lemmas that the proof relies upon should be properly referenced. 
    \end{itemize}

\item {\bf Experimental result reproducibility}
    \item[] Question: Does the paper fully disclose all the information needed to reproduce the main experimental results of the paper to the extent that it affects the main claims and/or conclusions of the paper (regardless of whether the code and data are provided or not)?
    \item[] Answer: \answerYes{}
    \item[] Justification: Sections~\ref{sec:method} and~\ref{sec:trace-analysis-auto} and Appendices~\ref{sec:data-collection} and~\ref{sec:additional-validation} describe corpus selection, codebook construction, the annotation pipeline, and agreement measurements. The released codebook, source corpora, analysis skills, and annotated trajectory examples provide a path to reproduce or verify the taxonomy and behavioral analyses.
    \item[] Guidelines:
    \begin{itemize}
        \item The answer \answerNA{} means that the paper does not include experiments.
        \item If the paper includes experiments, a \answerNo{} answer to this question will not be perceived well by the reviewers: Making the paper reproducible is important, regardless of whether the code and data are provided or not.
        \item If the contribution is a dataset and\slash or model, the authors should describe the steps taken to make their results reproducible or verifiable. 
        \item Depending on the contribution, reproducibility can be accomplished in various ways. For example, if the contribution is a novel architecture, describing the architecture fully might suffice, or if the contribution is a specific model and empirical evaluation, it may be necessary to either make it possible for others to replicate the model with the same dataset, or provide access to the model. In general. releasing code and data is often one good way to accomplish this, but reproducibility can also be provided via detailed instructions for how to replicate the results, access to a hosted model (e.g., in the case of a large language model), releasing of a model checkpoint, or other means that are appropriate to the research performed.
        \item While NeurIPS does not require releasing code, the conference does require all submissions to provide some reasonable avenue for reproducibility, which may depend on the nature of the contribution. For example
        \begin{enumerate}
            \item If the contribution is primarily a new algorithm, the paper should make it clear how to reproduce that algorithm.
            \item If the contribution is primarily a new model architecture, the paper should describe the architecture clearly and fully.
            \item If the contribution is a new model (e.g., a large language model), then there should either be a way to access this model for reproducing the results or a way to reproduce the model (e.g., with an open-source dataset or instructions for how to construct the dataset).
            \item We recognize that reproducibility may be tricky in some cases, in which case authors are welcome to describe the particular way they provide for reproducibility. In the case of closed-source models, it may be that access to the model is limited in some way (e.g., to registered users), but it should be possible for other researchers to have some path to reproducing or verifying the results.
        \end{enumerate}
    \end{itemize}

\item {\bf Open access to data and code}
    \item[] Question: Does the paper provide open access to the data and code, with sufficient instructions to faithfully reproduce the main experimental results, as described in supplemental material?
    \item[] Answer: \answerYes{}
    \item[] Justification: The public repository (\url{https://github.com/gaojie058/Act-onomy}) provides the codebook, corpora, analysis skills, usage instructions, and case-study examples; trajectory annotations are also linked on the title page. Appendices~\ref{sec:data-collection} and~\ref{sec:appendix:skill} describe corpus preparation and the trajectory-annotation pipeline.
    \item[] Guidelines:
    \begin{itemize}
        \item The answer \answerNA{} means that paper does not include experiments requiring code.
        \item Please see the NeurIPS code and data submission guidelines (\url{https://neurips.cc/public/guides/CodeSubmissionPolicy}) for more details.
        \item While we encourage the release of code and data, we understand that this might not be possible, so \answerNo{} is an acceptable answer. Papers cannot be rejected simply for not including code, unless this is central to the contribution (e.g., for a new open-source benchmark).
        \item The instructions should contain the exact command and environment needed to run to reproduce the results. See the NeurIPS code and data submission guidelines (\url{https://neurips.cc/public/guides/CodeSubmissionPolicy}) for more details.
        \item The authors should provide instructions on data access and preparation, including how to access the raw data, preprocessed data, intermediate data, and generated data, etc.
        \item The authors should provide scripts to reproduce all experimental results for the new proposed method and baselines. If only a subset of experiments are reproducible, they should state which ones are omitted from the script and why.
        \item At submission time, to preserve anonymity, the authors should release anonymized versions (if applicable).
        \item Providing as much information as possible in supplemental material (appended to the paper) is recommended, but including URLs to data and code is permitted.
    \end{itemize}

\item {\bf Experimental setting/details}
    \item[] Question: Does the paper specify all the training and test details (e.g., data splits, hyperparameters, how they were chosen, type of optimizer) necessary to understand the results?
    \item[] Answer: \answerYes{}
    \item[] Justification: There is no model training. Sections~\ref{sec:method} and~\ref{sec:trace-analysis-auto} and Appendices~\ref{sec:data-collection} and~\ref{sec:additional-validation} describe the corpus splits, annotation procedure, annotator models, evaluation samples, and agreement measures needed to interpret the results.
    \item[] Guidelines:
    \begin{itemize}
        \item The answer \answerNA{} means that the paper does not include experiments.
        \item The experimental setting should be presented in the core of the paper to a level of detail that is necessary to appreciate the results and make sense of them.
        \item The full details can be provided either with the code, in appendix, or as supplemental material.
    \end{itemize}

\item {\bf Experiment statistical significance}
    \item[] Question: Does the paper report error bars suitably and correctly defined or other appropriate information about the statistical significance of the experiments?
    \item[] Answer: \answerYes{}
    \item[] Justification: Section~\ref{sec:case-study-2} and Appendix~\ref{sec:pattern-validation} report 95\% confidence intervals for outcome comparisons, estimated with 10{,}000 trajectory-level bootstrap resamples. Appendix~\ref{sec:additional-validation} also reports confidence intervals for human--human agreement and significance tests for agreement in action rankings; some other agreement results are reported as point estimates.
    \item[] Guidelines:
    \begin{itemize}
        \item The answer \answerNA{} means that the paper does not include experiments.
        \item The authors should answer \answerYes{} if the results are accompanied by error bars, confidence intervals, or statistical significance tests, at least for the experiments that support the main claims of the paper.
        \item The factors of variability that the error bars are capturing should be clearly stated (for example, train/test split, initialization, random drawing of some parameter, or overall run with given experimental conditions).
        \item The method for calculating the error bars should be explained (closed form formula, call to a library function, bootstrap, etc.)
        \item The assumptions made should be given (e.g., Normally distributed errors).
        \item It should be clear whether the error bar is the standard deviation or the standard error of the mean.
        \item It is OK to report 1-sigma error bars, but one should state it. The authors should preferably report a 2-sigma error bar than state that they have a 96\% CI, if the hypothesis of Normality of errors is not verified.
        \item For asymmetric distributions, the authors should be careful not to show in tables or figures symmetric error bars that would yield results that are out of range (e.g., negative error rates).
        \item If error bars are reported in tables or plots, the authors should explain in the text how they were calculated and reference the corresponding figures or tables in the text.
    \end{itemize}

\item {\bf Experiments compute resources}
    \item[] Question: For each experiment, does the paper provide sufficient information on the computer resources (type of compute workers, memory, time of execution) needed to reproduce the experiments?
    \item[] Answer: \answerYes{}
    \item[] Justification: Non-LLM processing for the original experiments ran on a single Apple M-series laptop with 16\,GB RAM, no local GPU. LLM inference was provided through hosted services; Section~\ref{sec:trace-analysis-auto} and Appendix~\ref{sec:additional-validation} describe the annotation pipeline and annotator models.
    \item[] Guidelines:
    \begin{itemize}
        \item The answer \answerNA{} means that the paper does not include experiments.
        \item The paper should indicate the type of compute workers CPU or GPU, internal cluster, or cloud provider, including relevant memory and storage.
        \item The paper should provide the amount of compute required for each of the individual experimental runs as well as estimate the total compute. 
        \item The paper should disclose whether the full research project required more compute than the experiments reported in the paper (e.g., preliminary or failed experiments that didn't make it into the paper). 
    \end{itemize}
    
\item {\bf Code of ethics}
    \item[] Question: Does the research conducted in the paper conform, in every respect, with the NeurIPS Code of Ethics \url{https://neurips.cc/public/EthicsGuidelines}?
    \item[] Answer: \answerYes{}
    \item[] Justification: We have read the NeurIPS Code of Ethics. The reported work analyzes published text and agent trajectories, with co-author qualitative coding and no recruited external participants.
    \item[] Guidelines:
    \begin{itemize}
        \item The answer \answerNA{} means that the authors have not reviewed the NeurIPS Code of Ethics.
        \item If the authors answer \answerNo, they should explain the special circumstances that require a deviation from the Code of Ethics.
        \item The authors should make sure to preserve anonymity (e.g., if there is a special consideration due to laws or regulations in their jurisdiction).
    \end{itemize}

\item {\bf Broader impacts}
    \item[] Question: Does the paper discuss both potential positive societal impacts and negative societal impacts of the work performed?
    \item[] Answer: \answerYes{}
    \item[] Justification: Appendix~\ref{sec:impacts} discusses the benefits of a shared vocabulary for describing agent behavior, the risk of treating behavioral profiles as ground truth, constraints imposed by a shared vocabulary, under-represented literature, and dependence on commercial models.
    \item[] Guidelines:
    \begin{itemize}
        \item The answer \answerNA{} means that there is no societal impact of the work performed.
        \item If the authors answer \answerNA{} or \answerNo, they should explain why their work has no societal impact or why the paper does not address societal impact.
        \item Examples of negative societal impacts include potential malicious or unintended uses (e.g., disinformation, generating fake profiles, surveillance), fairness considerations (e.g., deployment of technologies that could make decisions that unfairly impact specific groups), privacy considerations, and security considerations.
        \item The conference expects that many papers will be foundational research and not tied to particular applications, let alone deployments. However, if there is a direct path to any negative applications, the authors should point it out. For example, it is legitimate to point out that an improvement in the quality of generative models could be used to generate Deepfakes for disinformation. On the other hand, it is not needed to point out that a generic algorithm for optimizing neural networks could enable people to train models that generate Deepfakes faster.
        \item The authors should consider possible harms that could arise when the technology is being used as intended and functioning correctly, harms that could arise when the technology is being used as intended but gives incorrect results, and harms following from (intentional or unintentional) misuse of the technology.
        \item If there are negative societal impacts, the authors could also discuss possible mitigation strategies (e.g., gated release of models, providing defenses in addition to attacks, mechanisms for monitoring misuse, mechanisms to monitor how a system learns from feedback over time, improving the efficiency and accessibility of ML).
    \end{itemize}
    
\item {\bf Safeguards}
    \item[] Question: Does the paper describe safeguards that have been put in place for responsible release of data or models that have a high risk for misuse (e.g., pre-trained language models, image generators, or scraped datasets)?
    \item[] Answer: \answerNA{}
    \item[] Justification: We release a codebook, a prompt template, and labeled examples. We do not release pretrained models, image generators, or web-scraped datasets; the materials carry no foreseeable high-misuse risk that warrants gated release.
    \item[] Guidelines:
    \begin{itemize}
        \item The answer \answerNA{} means that the paper poses no such risks.
        \item Released models that have a high risk for misuse or dual-use should be released with necessary safeguards to allow for controlled use of the model, for example by requiring that users adhere to usage guidelines or restrictions to access the model or implementing safety filters. 
        \item Datasets that have been scraped from the Internet could pose safety risks. The authors should describe how they avoided releasing unsafe images.
        \item We recognize that providing effective safeguards is challenging, and many papers do not require this, but we encourage authors to take this into account and make a best faith effort.
    \end{itemize}

\item {\bf Licenses for existing assets}
    \item[] Question: Are the creators or original owners of assets (e.g., code, data, models), used in the paper, properly credited and are the license and terms of use explicitly mentioned and properly respected?
    \item[] Answer: \answerNo{}
    \item[] Justification: The original papers and trajectory sources are credited in Sections~\ref{sec:method} and~\ref{sec:case-study} and Appendix~\ref{sec:raw-math-trajectories}. However, the paper does not explicitly enumerate the licenses and terms of use for every third-party dataset, codebase, and model service.
    \item[] Guidelines:
    \begin{itemize}
        \item The answer \answerNA{} means that the paper does not use existing assets.
        \item The authors should cite the original paper that produced the code package or dataset.
        \item The authors should state which version of the asset is used and, if possible, include a URL.
        \item The name of the license (e.g., CC-BY 4.0) should be included for each asset.
        \item For scraped data from a particular source (e.g., website), the copyright and terms of service of that source should be provided.
        \item If assets are released, the license, copyright information, and terms of use in the package should be provided. For popular datasets, \url{paperswithcode.com/datasets} has curated licenses for some datasets. Their licensing guide can help determine the license of a dataset.
        \item For existing datasets that are re-packaged, both the original license and the license of the derived asset (if it has changed) should be provided.
        \item If this information is not available online, the authors are encouraged to reach out to the asset's creators.
    \end{itemize}

\item {\bf New assets}
    \item[] Question: Are new assets introduced in the paper well documented and is the documentation provided alongside the assets?
    \item[] Answer: \answerYes{}
    \item[] Justification: Appendix~\ref{sec:full-codebook} documents the taxonomy with definitions and quoted examples, and the public repository provides the codebook, analysis skills, and labeled case-study examples. The repository specifies the MIT License for source code and CC BY 4.0 for codebook content and labeled datasets.
    \item[] Guidelines:
    \begin{itemize}
        \item The answer \answerNA{} means that the paper does not release new assets.
        \item Researchers should communicate the details of the dataset\slash code\slash model as part of their submissions via structured templates. This includes details about training, license, limitations, etc. 
        \item The paper should discuss whether and how consent was obtained from people whose asset is used.
        \item At submission time, remember to anonymize your assets (if applicable). You can either create an anonymized URL or include an anonymized zip file.
    \end{itemize}

\item {\bf Crowdsourcing and research with human subjects}
    \item[] Question: For crowdsourcing experiments and research with human subjects, does the paper include the full text of instructions given to participants and screenshots, if applicable, as well as details about compensation (if any)? 
    \item[] Answer: \answerNA{}
    \item[] Justification: The qualitative coding in the reported studies was performed by the paper's co-authors as part of normal research activity. No crowdworkers or external participants were recruited for these analyses, so compensation and participant-instruction disclosures do not apply.
    \item[] Guidelines:
    \begin{itemize}
        \item The answer \answerNA{} means that the paper does not involve crowdsourcing nor research with human subjects.
        \item Including this information in the supplemental material is fine, but if the main contribution of the paper involves human subjects, then as much detail as possible should be included in the main paper. 
        \item According to the NeurIPS Code of Ethics, workers involved in data collection, curation, or other labor should be paid at least the minimum wage in the country of the data collector. 
    \end{itemize}

\item {\bf Institutional review board (IRB) approvals or equivalent for research with human subjects}
    \item[] Question: Does the paper describe potential risks incurred by study participants, whether such risks were disclosed to the subjects, and whether Institutional Review Board (IRB) approvals (or an equivalent approval/review based on the requirements of your country or institution) were obtained?
    \item[] Answer: \answerNA{}
    \item[] Justification: The reported analyses involve published text, agent trajectories, and co-author qualitative coding, with no recruited external human participants.
    \item[] Guidelines:
    \begin{itemize}
        \item The answer \answerNA{} means that the paper does not involve crowdsourcing nor research with human subjects.
        \item Depending on the country in which research is conducted, IRB approval (or equivalent) may be required for any human subjects research. If you obtained IRB approval, you should clearly state this in the paper. 
        \item We recognize that the procedures for this may vary significantly between institutions and locations, and we expect authors to adhere to the NeurIPS Code of Ethics and the guidelines for their institution. 
        \item For initial submissions, do not include any information that would break anonymity (if applicable), such as the institution conducting the review.
    \end{itemize}

\item {\bf Declaration of LLM usage}
    \item[] Question: Does the paper describe the usage of LLMs if it is an important, original, or non-standard component of the core methods in this research? Note that if the LLM is used only for writing, editing, or formatting purposes and does \emph{not} impact the core methodology, scientific rigor, or originality of the research, declaration is not required.
    \item[] Answer: \answerYes{}
    \item[] Justification: Section~\ref{sec:method-phase1} and Appendix~\ref{app:discovery-judge} describe LLM-assisted behavior extraction and codebook expansion under human review. Section~\ref{sec:trace-analysis-auto} and Appendix~\ref{sec:additional-validation} describe LLM-based trajectory annotation and comparisons across models and human coders; LLMs were also used for routine editorial assistance.
    \item[] Guidelines:
    \begin{itemize}
        \item The answer \answerNA{} means that the core method development in this research does not involve LLMs as any important, original, or non-standard components.
        \item Please refer to our LLM policy in the NeurIPS handbook for what should or should not be described.
    \end{itemize}

\end{enumerate}

\end{document}